%% file: main-v1.tex
\documentclass{article}

\PassOptionsToPackage{numbers, compress}{natbib}

\usepackage[preprint]{neurips_2021}

\usepackage[utf8]{inputenc} 
\usepackage[T1]{fontenc}    
\usepackage{hyperref}       
\usepackage{url}            
\usepackage{booktabs}       
\usepackage{amsfonts}       
\usepackage{nicefrac}       
\usepackage{microtype}      
\usepackage{xcolor}         
\usepackage{subcaption}
\usepackage{multirow}

\usepackage{adjustbox}

\usepackage{graphicx}
\usepackage{caption}

\usepackage{bm,amsmath}
\usepackage{bbm}
\usepackage{sidecap}
\usepackage{tikz}
\usetikzlibrary{shapes,arrows}
\tikzstyle{block} = [draw, fill=blue!20, rectangle, minimum height=3em, minimum width=6em]
\tikzstyle{sum} = [draw, fill=blue!20, circle, node distance=1cm]
\tikzstyle{input} = [coordinate]
\tikzstyle{output} = [coordinate]
\tikzstyle{pinstyle} = [pin edge={to-,thin,black}]

\definecolor{darkred}{rgb}{1, 0.1, 0.3}
\definecolor{darkblue}{rgb}{0.1, 0.1, 1}
\definecolor{darkgreen}{rgb}{0,0.6,0.5}

\newcommand {\mm}[1] {\ifmmode{#1}\else{\mbox{\(#1\)}}\fi}

\hypersetup{colorlinks=true,allcolors=[rgb]{0,0.25,0.65}}

\def \proposed {\text{SAND}} 
\input{math_commands.tex}

\title{SAND-mask: An Enhanced Gradient Masking Strategy for the Discovery of Invariances in Domain Generalization}

\author{%
Soroosh Shahtalebi$^{1~2}$\\
\And
Jean-Christophe Gagnon-Audet$^{1~2}$\\
\And
Touraj Laleh$^{1~2}$\\
\And
Mojtaba Faramarzi$^{1~2}$\\
\And
Kartik Ahuja$^{1~2}$\\
\And
Irina Rish$^{1~2}$\\
\And
\\
$^{1}$Mila - Quebec AI Institute, Canada \\ $^{2}$ Université de Montréal, Département d’Informatique et Recherche Opérationelle, Montreal, Canada\\
Correspondence to: <\texttt{soroosh.shahtalebi@mila.quebec}> \\
}

\begin{document}

\maketitle

\begin{abstract}
A major bottleneck in the real-world applications of machine learning models is their failure in generalizing to unseen domains whose data distribution is not \textit{i.i.d} to the training domains. This failure often stems from learning non-generalizable features in the training domains that are spuriously correlated with the label of data. To address this shortcoming, there has been a growing surge of interest in \textit{learning good explanations that are hard to vary}, which is studied under the notion of \textit{Out-of-Distribution (OOD) Generalization}. The search for good explanations that are \textit{invariant} across different domains can be seen as finding local (global) minimas in the loss landscape that hold true across all of the training domains. In this paper, we propose a masking strategy, which determines a continuous weight based on the \textit{agreement} of gradients that flow in each edge of network, in order to control the amount of update received by the edge in each step of optimization. Particularly, our proposed technique referred to as ``Smoothed-AND (SAND)-masking'', not only validates the agreement in the direction of gradients but also promotes the agreement among their magnitudes to further ensure the discovery of invariances across training domains. SAND-mask is validated over the Domainbed benchmark for domain generalization and significantly improves the state-of-the-art accuracy on the Colored MNIST dataset, while providing competitive results on other domain generalization datasets.

\end{abstract}
\section{Introduction}\label{sec:intro}

Although machine learning models have shown promising performance in various different applications such as computer vision~\cite{he2016deep,krizhevsky2012imagenet}, speech recognition~\cite{graves2013speech}, and natural language processing~\cite{devlin2018bert}, they often fail to generalize beyond the training distribution. In other words, conventional machine learning techniques assume that test data is \textit{i.i.d} with respect to the training data, which is often violated in practical applications. Examples of this failure mode include adversarial attacks~\cite{madry2017towards}, spurious correlations~\cite{irm}, population shifts~\cite{koh2020wilds}, and naturally-occurred variations in the distribution of data~\cite{robey2020model}. To mitigate this shortcoming, there has been a growing surge of interest in learning efficient cues in the training data, which hold true across unseen domains. This topic of research, which is referred to as ``\textit{Domain Generalization}'' or ``\textit{Out-of-distribution (OOD) Generalization}'', particularly aims at recognizing and penalizing the features that are spuriously correlated with label, thus facilitating the learning of ``\textit{good features}'' that are assumed to generalize out of domain.

The search for consistencies across domains often occurs at the feature level~\cite{irm,decaug,gradientstarvation,iga} or at the gradients level~\cite{ilc,shi2021gradient}, where the former aims at generating latent variables that efficiently represent all the training domains and thus minimize the risk across all of the domains. The latter, on the other hand, aims at promoting the agreement among the gradients form training examples form different domains. This sort of agreement is either enforced through regularizer terms in the objective function~\cite{shi2021gradient} or is just encouraged by masking the gradients that point to the same direction~\cite{ilc}. It is worth noting that despite the considerable growth in the body of literature on OOD generalization,~\citet{domainbed} empirically shown that non of the existing works significantly outperforms the classical Empirical Risk Minimization (ERM) objective for training the learning models under a setting where algorithms doesn't have access to a test set to finetune hyperparameters. This further corroborates the urge for efficient methodologies to distinguish between spurious and causal (invariant) features, which will provide the capability to generalize out of distribution.

The search for consistencies among different domains to train a model solely based on invariant explanations is often fulfilled by acquiring training samples from different domains, which on the other hand, demands for excessive training data. Recently, \citet{ilc} proposed a new strategy, referred to as \textit{``Invariant Learning Consistency (ILC)''} in the search for consistencies among different domains, which treats each data sample as a separate domain, and thus aims at finding invariances among them. This strategy, which is fulfilled by only backpropagating the gradients from a batch of data that consistently point to a certain direction, aims at promoting the parts of Hessians that different domains (environments) agree the most, while mitigating the need for training data from several domains. In practice, this strategy takes the form of a discrete mask, called "AND-mask", which is applied to the gradients. Although effective in some curated test conditions, their proposed mechanism to check and promote the agreement among the direction of gradients suffers from a number of failure modes, such as reliance on the momentum term in optimizer, susceptibility to initialization, and susceptibility to noise on training data (thoroughly discussed in Subsection~\ref{subsec:failmodes}). This paper addresses the shortcomings of ILC method. 

Additionally, some recent works in OOD generalization, including Invariant Risk Minimization (IRM)~\cite{irm}, Risk Extrapolation (REx)~\cite{rex} and Spectral Decoupling~\cite{gradientstarvation} have introduced and employed an annealing parameter. It determines the step in the training procedure that the OOD generalization technique should become the main focus of the objective. In the annealing strategy, the training starts with a small penalty weight in order to allow the model to learn predictive feature and after the annealing step the penalty weight is sharply increased so the model can select the invariant feature from those predictive feature. It is hypothesized that this strategy helps the network to leave the initialization point and find its way towards to the optimal solution. Although this strategy is empirically shown to improve the test-time accuracy, there is currently no robust theoretical or empirical rationale on how to identify this switching threshold without access to a representative test set which is impossible in a true OOD setting. Motivated by the promising effect of this strategy~\cite{irm,rex}, our proposed methodology automatically converges from no masking to our proposed $\proposed$ mask, based on the agreement among the gradients in each edge of the network. In other words, $\proposed$ mask does not impose a certain annealing parameter for all the parameters of a network, and changes its shape based on the agreement of gradients flowing in each edge of network. The transition is not only based on the agreement among the direction gradients but also depends on the agreement among the magnitude of gradients. Therefore, our proposed SAND-mask favors matching the Hessians of different environments by simultaneously checking the agreement among the magnitude and direction of the backpropagated gradients. In summary, the paper offers the following contributions:
\begin{itemize}
\item The proposed masking strategy, SAND-mask, not only takes the direction of the gradients into account, but also values the agreement between the magnitude of gradients in order to match the Hessians of different environments and ensure learning invariant features across the training domains. 
\item Unlike conventional methods in OOD Generalization, which define a handcrafted criteria to turn on their proposed methodology (regularizer, objective function, or gradient operation), the proposed $\proposed$-masking strategy automatically and individually for each parameter of the model changes its shape based on the level of agreement in the gradients.  
\end{itemize}

\section{Related Works}\label{sec:relworks}
As the name of paper suggests, this work aims at devising a methodology to train an invariant predictor that helps with generalizing to out of domain distributions, i.e., domain generalization. However, throughout the paper, the terms ``domain generalization'' and ``OOD generalization'' are used interchangeably as if they carry the same meaning. To clarify this, it is worth noting that OOD generalization refers to the case that the test distribution is different from the training one, which does not take into account any notion of domain. On the other hand, domain generalization concerns the case that a model is being tested over a domain, which is never seen in the training phase. Since the collected training samples from each domain are assumed to completely represent the data distribution in that domain (the entire range of variations in the features), we believe that the two problems of domain generalization and OOD generalization would become the two sides of the same coin.

Considering the methodologies employed to disentangle spurious features from the invariant ones, the body of knowledge on OOD generalization can be categorized into two groups; (1) techniques that enforce/encourage agreement at representation level, and (2) techniques that enforce/encourage agreement in gradient level. Here, we briefly review the literature on these two approaches. 

\begin{enumerate}

\item \textbf{Representation-level agreement:}
This approach in domain generalization, which is extensively studied in the literature, aims at training a model that treats samples with the same label but from different domains as the same and yields similar representations for them. In other words, the goal here is to learn models that map different domains (different distributions) into a single statistical distributions~\cite{irm}. A trivial approach to fulfill this goal is to match the mean and variance of the representations across domains~\cite{sun2016deep} or to match the distribution of representations~\cite{li2018domain}. Another approach is to penalize the domain-predictive power of the representations in order to achieve indistinguishable representations for the training domains. In addition, the representation-level agreement can also be satisfied by comparing and minimizing the average risk (ERM) or the maximum risk for the training domains~\cite{rojas2015causal}. It is worth noting that ERM technique still offers the best OOD generalization performance on many datasets~\cite{domainbed,koh2020wilds}.

\item \textbf{Gradient-level agreement:} This approach, as opposed to the previous one, aims at finding local or global minimas in the loss space that are common across all of the training domains. In other words, the goal here is to have the network to share similar Hessians for different domains, and this is often fulfilled by studying the gradients that are backpropagated in the network. To this goal, the work in~\cite{iga} aims at minimizing the variance of inter-domain gradients to enhance the agreement (alignment) between the gradients. In addition, the work in~\cite{shi2021gradient} measures the alignment of inter-domain gradients by computing their inner product, and then penalizes the network such that the dispersion of gradients gets minimized. Finally, the work in~\cite{ilc} proposes an AND-masking strategy which checks the agreement in the direction to which gradients are pointing and allows the parameters of network to be update only if all the gradients flowing in that parameter agree on a certain direction. This strict masking strategy is analogous to applying a logical AND operator on the direction of gradients. It is worth noting that while the works in~\cite{iga} and~\cite{shi2021gradient} enforce the maximal alignment of gradients by including a regularizing term in the objective function of model, the work in~\cite{ilc} just encourages the alignment by filtering out the gradients that point to different directions in the loss space.

\end{enumerate}

\section{Problem Formulation}\label{sec:probform}
In this section, a formal definition and description of the problem in hand is provided. Since our proposed SAND-masking technique serves as an extension to the AND-masking technique, in this work and for consistency, we follow the same style and notations that are used in the work of~\citet{ilc}. 

\subsection{Invariant Leaning Consistency; AND-masking}
Assuming that we have $\{\mathcal{D}^e = (x_i^e, y_i^e) \}_{e \in \mathcal{E}}$ datasets, where $e$ is the superscript for the environment from which data is collected, $i_e = 1,\ldots,n^e$, and $|\mathcal{E}| = d$ is the number of environments. Also, $x_i^e \in \mathcal{X} \subseteq \mathbb{R}^m$ denotes the vector of observed data, and $y_i^e \in \mathcal{Y} \subseteq \mathbb{R}^p$ is the vector of labels associated with the inputs. The goal is to learn a function (mechanism) $f: \mathcal{X} \xrightarrow[]{} \mathcal{Y}$ that captures the invariant features across different environments and thus provides a reusable mechanism to be used on unseen environments. In this work, the function $f$ is approximated by a neural network with parameters $\theta \in \mathcal{\theta} \subseteq \mathbb{R}^n$, and the output of the neural network is denoted by $f_\theta(x)$.

As discussed earlier, the approach taken to capture the consistencies of several domains in ILC technique as well as this work is to compare the Hessians of environments and locate some regions where the landscapes looks similar to each other~\cite{ilc}. However, the arithmetic averaging of Hessians might fail to capture the inconsistencies of landscapes due to the bias that might be induced by some environments with dominant features. Therefore,~\citet{ilc} proposed \textit{geometric averaging} of Hessians as a means for capturing the consistencies of environments. As opposed to arithmetic mean that performs a ``logical OR'' on the Hessians, geometric mean acts as a ``Logical AND'' operator and requires full consistency among environments~\cite{ando2004geometric}.

Assuming that the Hessian matrix of each environment, $H_e$, is diagonal~\cite{adolphs2019ellipsoidal,singh2020woodfisher} with positive eigenvalues, $\lambda_i^e$, the geometric mean of Hessians is $H^{\wedge} := \text{diag} ((\Pi_{e \in \mathcal{E}}\lambda_1^e)^{\frac{1}{|\mathcal{E}|}}, \ldots, (\Pi_{e \in \mathcal{E}}\lambda_n^e)^{\frac{1}{|\mathcal{E}|}})$. On the other hand, the arithmetic mean can be calculated as $H^{+} := \text{diag} ({\frac{1}{|\mathcal{E}|}\sum_{e\in\mathcal{E}}\lambda_1^e}, \ldots, {\frac{1}{|\mathcal{E}|}\sum_{e\in\mathcal{E}}\lambda_n^e})$. We recall that the conventional gradient descent method is based on the arithmetic average of Hessians for all the training environments and is calculated as $\theta^{k+1} = \theta^k - \eta H^+(\theta^k - \theta^*)$. In other words, the full gradient of network is $\nabla \mathcal{L}(\theta) = H^+ (\theta^k - \theta^*)$, which with the availability of geometric mean of Hessians can be rewritten as $\nabla \mathcal{L}^{\wedge}(\theta) = H^{\wedge} (\theta^k - \theta^*)$. Based on the definition of geometric mean, we can have $\nabla\mathcal{L}^{\wedge}(\theta) = (\Pi_{e \in \mathcal{E}} \nabla\mathcal{L}_e(\theta))^{\frac{1}{|\mathcal{E}|}}$, which means that the geometric mean of Hessians could be achieved by calculating the geometric average of element-wise gradients. However, to apply geometric averaging on gradients, it is crucially important that all the elements be consistent in the sign (direction) of gradient. This condition is validated by constructing and applying an AND-mask on the direction of gradients for each parameter of network. The AND-mask constructs a binary matrix $m_\tau(\theta^k)$ based on the agreement of direction of gradients and returns ``$1$'' if all the environments agree on a certain direction and ``$0$'' if otherwise. In other words, the mask for parameter $j$ of network is constructed as $[ m_{\tau} ]_j = \mathbbm{1}[\tau d \leq |\sum_e \text{sign}([ \nabla\mathcal{L}_e]_j)|]$, where $\tau$ is an agreement threshold $\tau \in [0,1]$ that identifies the portion of environments that need be agree, and $d$ is the number of environments. Finally, the mask is applied on gradients as in $m_\tau(\theta^k) \odot \nabla\mathcal{L} (\theta^k)$ that controls which parameters should receive updates based on the agreement of direction among the gradients flowing in that parameter.  

Despite the promising performance of the AND-masking technique~\cite{ilc}, we have identified a number of failure modes that limit its widespread and reliable application in different OOD generalization tasks. In what follows, a detailed list of failure modes and their effect on learning the invariant features is provided. 

\subsection{Failure Modes of AND-masking}\label{subsec:failmodes}

As discussed earlier, the AND-masking technique requires that the direction to which the gradients from different environments point strictly match with each other in order to allow the pooled gradients update that particular parameter. More formally, the direction of gradients for each component across $(\tau\times100) \%$ of environments must agree in order to have that component updated. It is evident that $\tau=1$ resembles the logical AND between the directions and $\tau=0$ is in fact the logical OR of the gradients. Assuming that the training data from $n_e$ number of environments is available, and taking into account that each gradient spans an infinite range in $\mathbb{R}$, the collection of gradients flowing into each parameter from all of the environments forms a $n_e-$dimensional space, which constitutes of $2^{n_e}$ orthants. Employing the AND-masking strategy, i.e. logical ANDing the gradients, technically means that only $2$ orthants of this space (the non-positive and the non-negative ones) would fulfill the AND-masking condition. In other words, by having $\tau=1$, there is a high probability ($\frac{2^{n_e}-2}{2^{n_e}}$) that each component get stuck in a neutral region, which we refer to as ``\textit{dead zone}'', and does not receive any update. It should be noted that selecting the agreement threshold imposes a tradeoff between the desirable characteristics of AND-masking and the number of dead zones in the loss space. 



\begin{figure}[t]
    \centering
    \subfloat[][Environment A]{\includegraphics[width=0.24\linewidth]{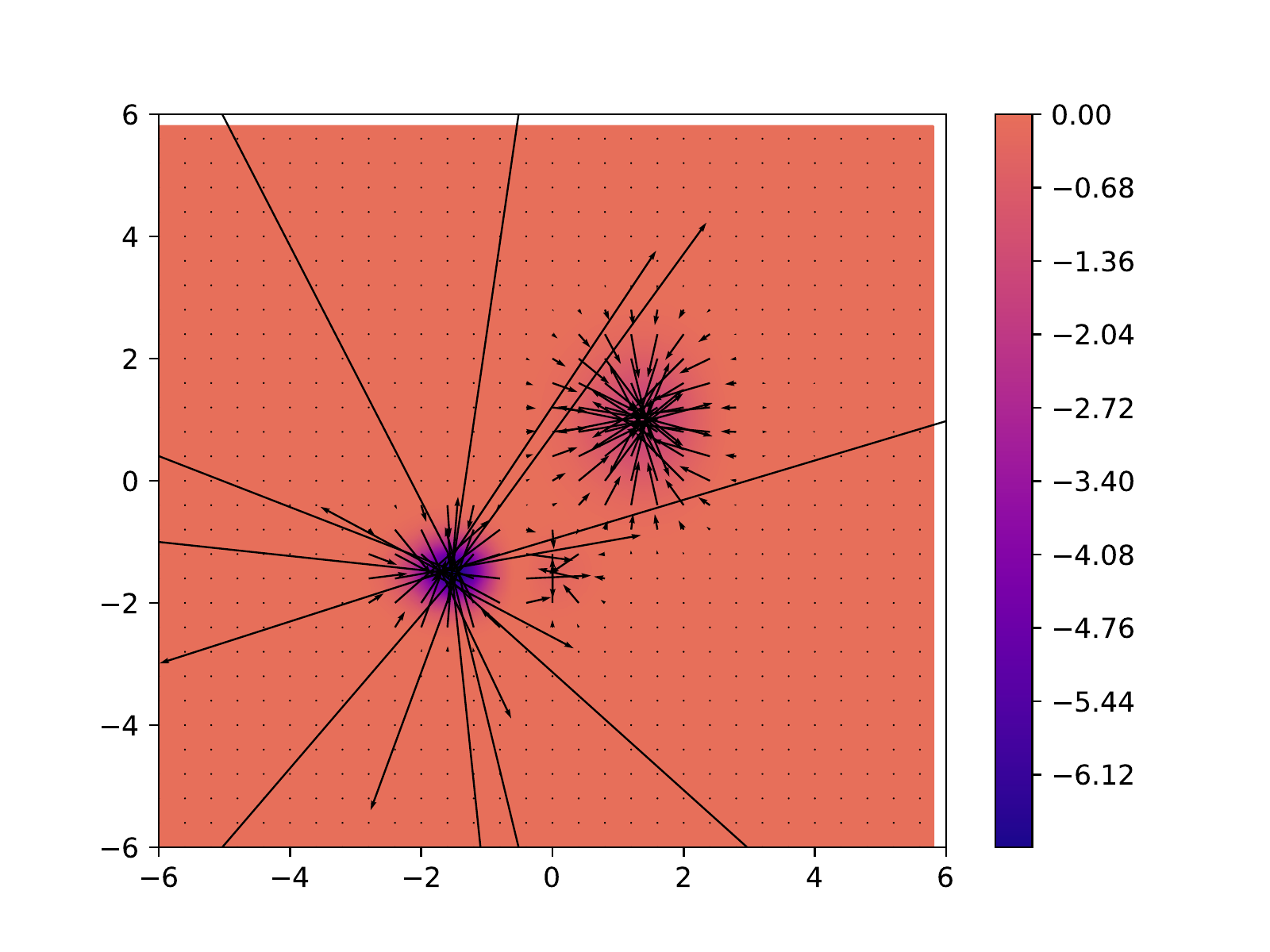}}
    \subfloat[][Arithmetic Avg]{\includegraphics[width=0.24\linewidth]{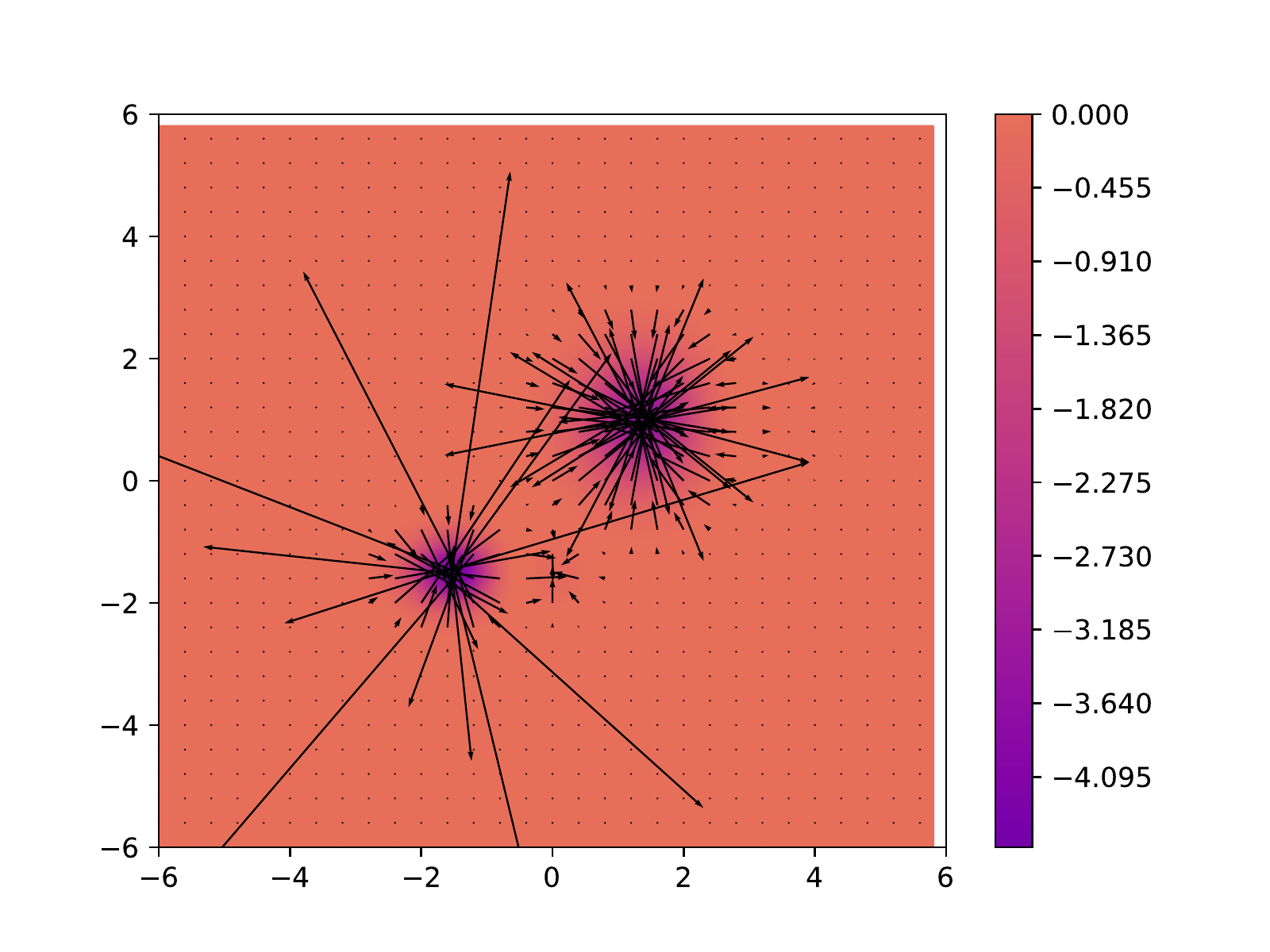}}
    \subfloat[][AND-mask (quiver)]{\includegraphics[width=0.24\linewidth]{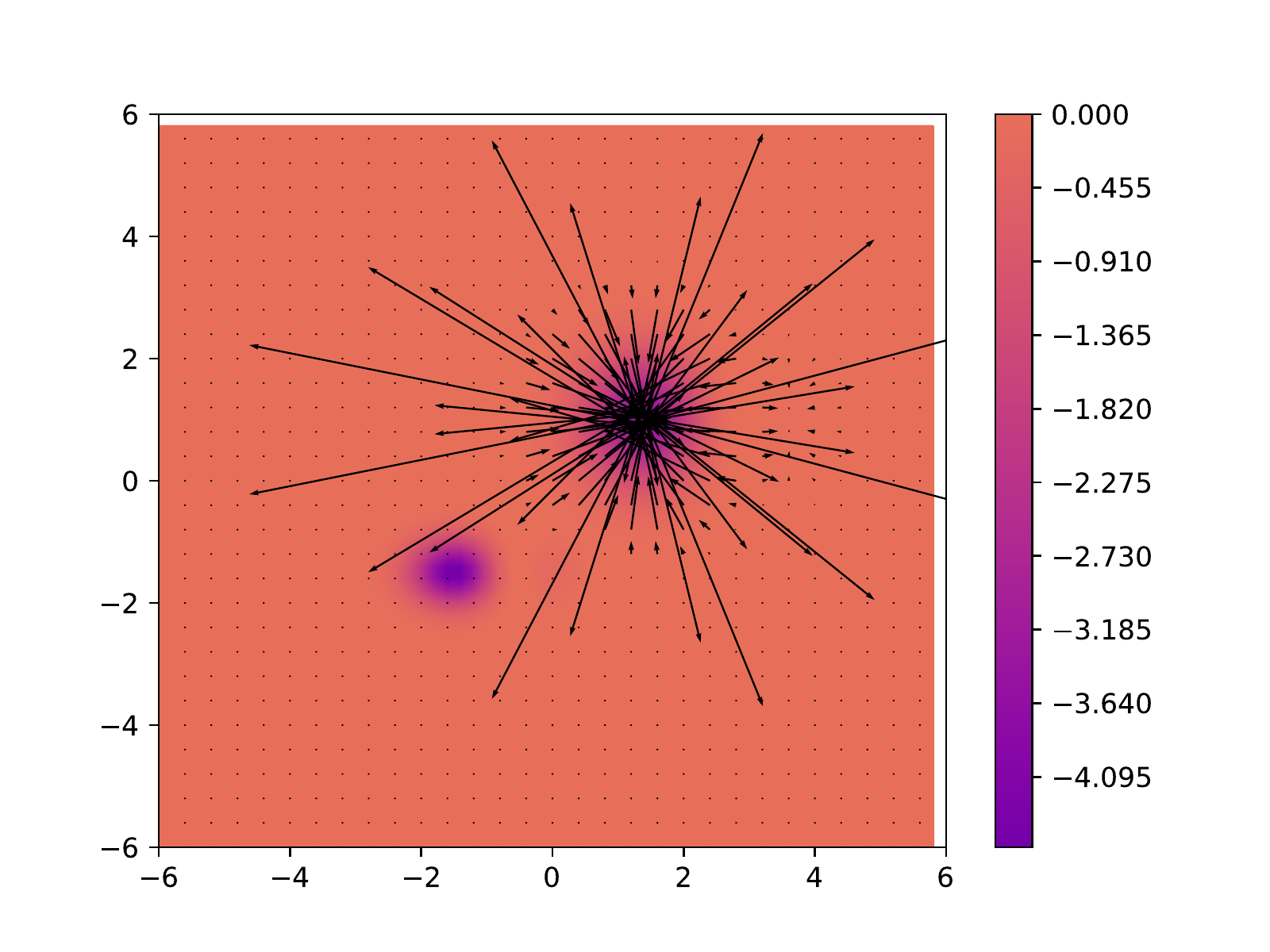}}
    \subfloat[][SAND-mask (quiver)]{\includegraphics[width=0.24\linewidth]{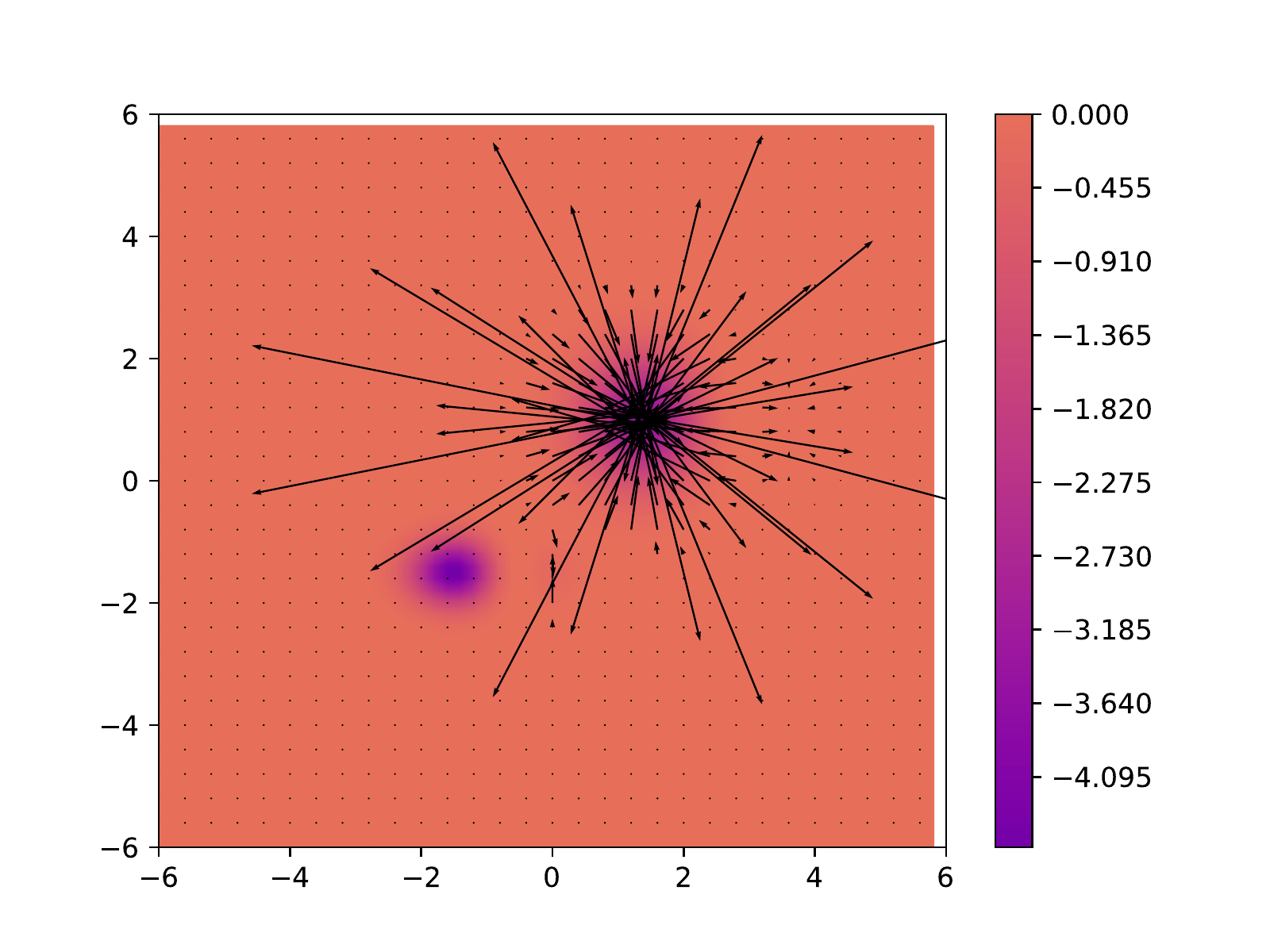}}\\
    \subfloat[][Environment B]{\includegraphics[width=0.25\linewidth]{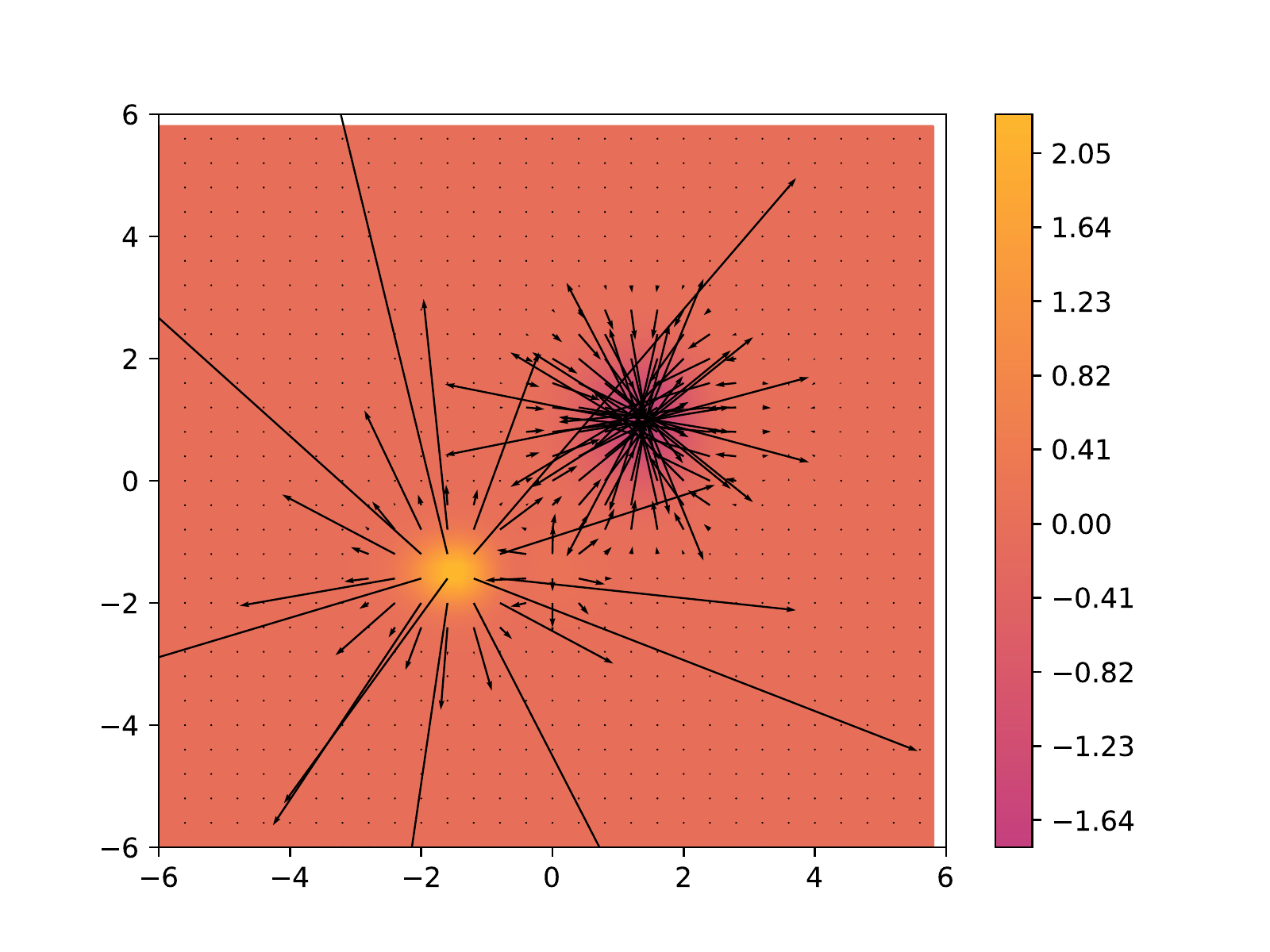}}\quad\quad\quad\quad\quad\quad\quad\quad
    \subfloat[][AND-mask (stream)]{\includegraphics[width=0.25\linewidth]{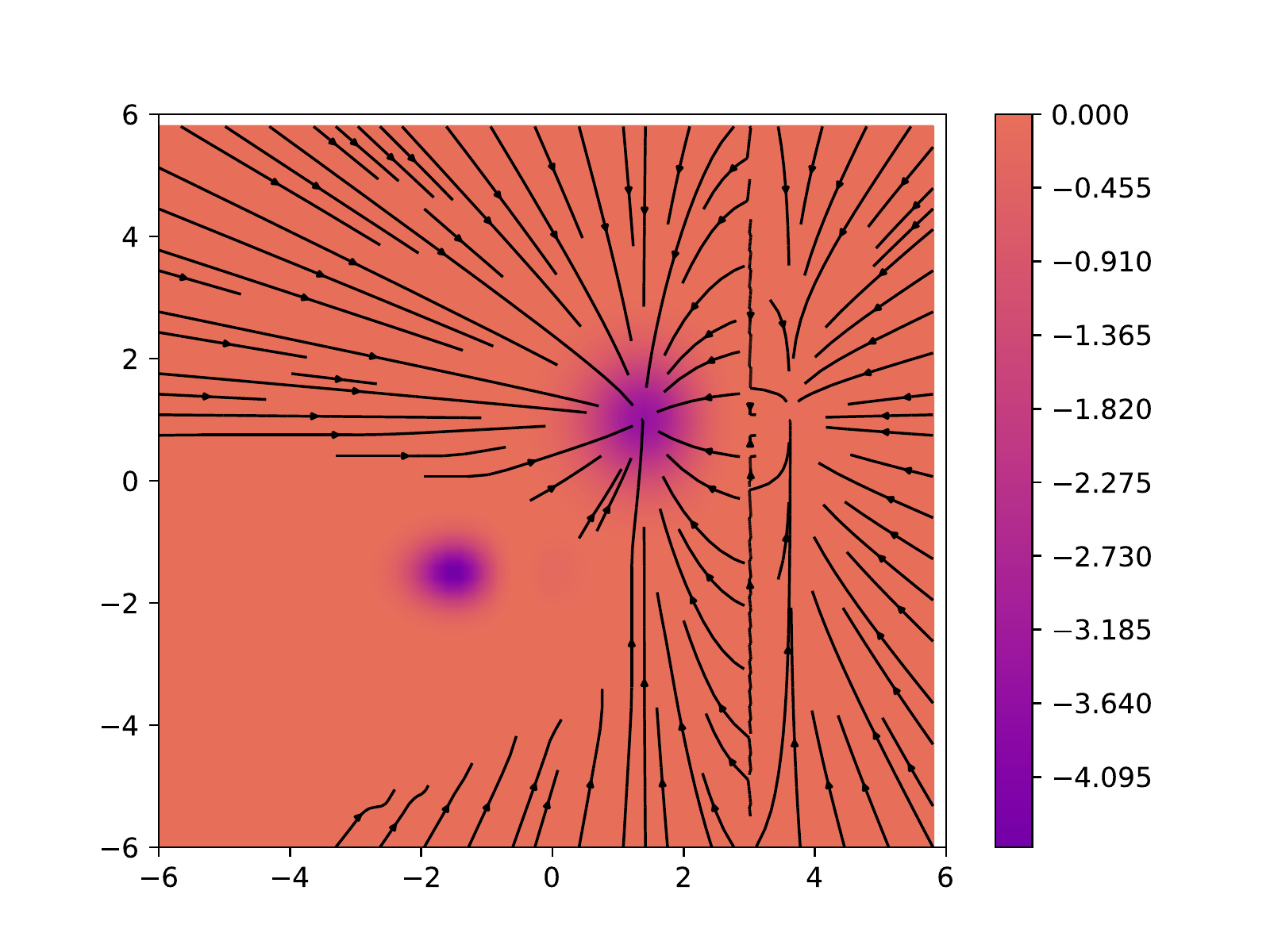}}
    \subfloat[][SAND-mask (stream)]{\includegraphics[width=0.25\linewidth]{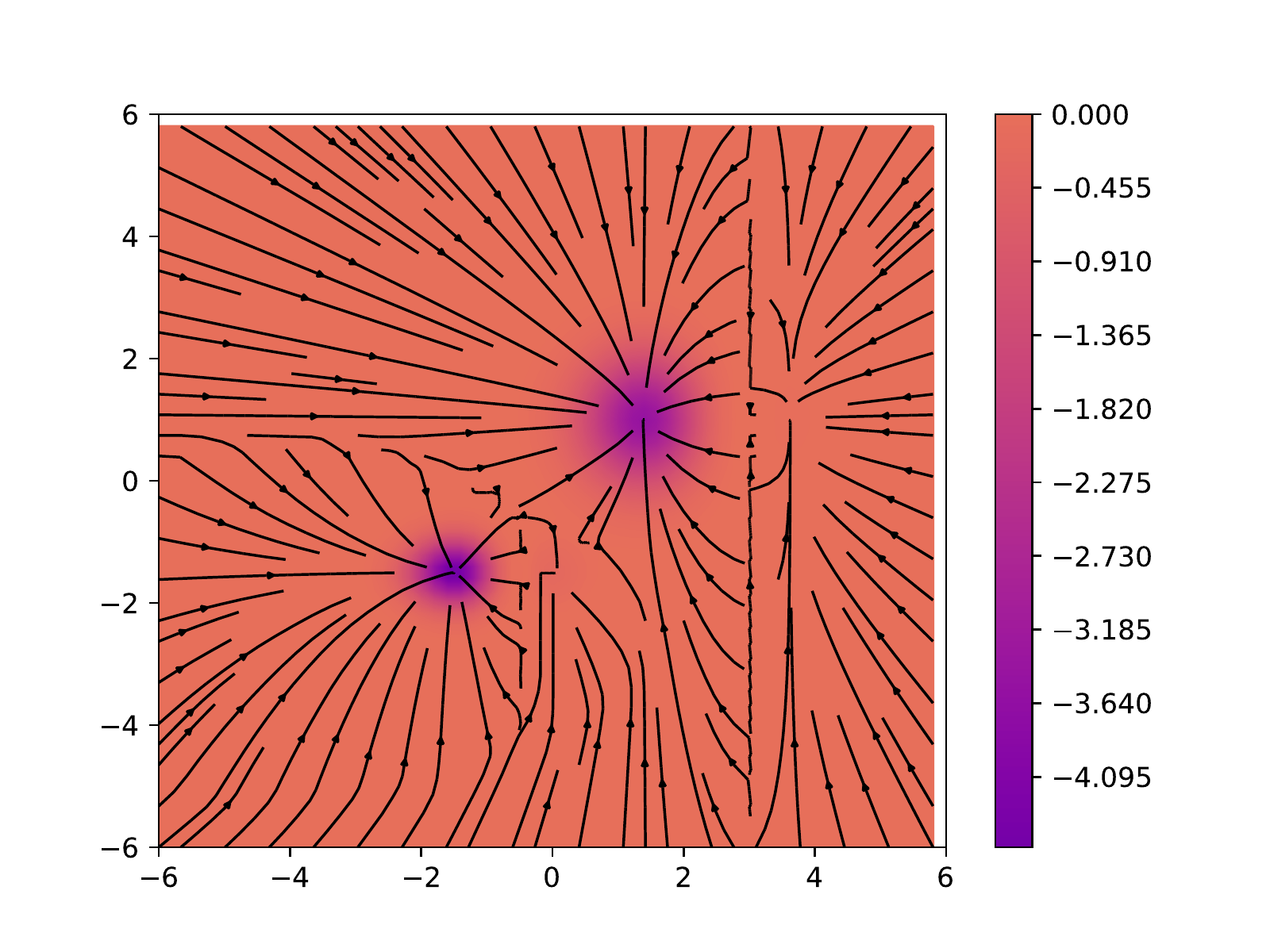}}\\
    \caption{(a) and (e) The loss landscape of two environments in our toy example; (b) The quiver plot of gradients over the arithmetic average of the loss landscapes; (c) and (f) The effect of AND-masking on the average loss landscape of two environments, which leads to creation of dead zones; (d) and (g) The effect of SAND-mask in avoiding the dead zones while providing the same average landscape for the two environments.}
    \label{fig:fail-landscape}
\end{figure}

To illustrate the above hypothesis regarding the role of AND-masking in formation of dead zones in loss landscape, a toy example inspired from the motivating example provided by~\citet{ilc} is formed. To this aim, we generated the loss landscapes of two environments in $2-$dimensional space, where both share a small and shallow local minimum on top-right of the origin of their landscapes. In addition, Environment A has a deep (global) minimum on bottom-left of its origin, while Environment B has a shallow local maximum on bottom-left of its origin (see Fig.~\ref{fig:fail-landscape}~(a) and~(e)). The depth of the inconsistent extremum on bottom-left of the two environments is selected such that the arithmetic average of loss surfaces reveals a local minimum in that region, as shown in Fig.~\ref{fig:fail-landscape}~(c). The arithmetic average of surfaces creates a local minimum on bottom-left of the origin, which although is in contrast to the consistency principle as declared by ILC, is indeed the point where the ERM technique identifies as one local minimum of the two surfaces. In other words, this toy example demonstrates a case where ERM will definitely fail in identifying the invariant features across the two domains. AND-mask and SAND-mask, which both follow the same principle in measuring the consistency, now need to be examined over this toy example. Please note that the heatmap in the background of Figs.~\ref{fig:fail-landscape}~(c),~(d),~(f) and~(g) represents the arithmetic average of the two loss landscapes. As it is shown in Figs.~\ref{fig:fail-landscape}~(c)~and~(d), both AND-mask and SAND-mask have selected the right gradients that preserve the invariant minima across the two environments. However, by considering the projection of gradients (stream plots), as it is observed in Fig.~\ref{fig:fail-landscape}~(f), the AND-masking technique contributes to the formation of a huge region in the average landscape that no trace of gradients can be found, which is indeed the dead zones that we theoretically detected. However, Fig.~\ref{fig:fail-landscape}~(d) clearly shows how the proposed SAND-mask alleviates the problem with the dead zones and allows the gradients to explore the whole landscape, and yet converge to the invariant feature.   


The presence of dead zones in the loss space gives rise to several problems, as listed below, which we group them together under the term ``failure modes of AND-masking''.

\paragraph{Heavy reliance on Momentum} The reason that AND-masking despite the presence of dead zones in the loss landscape can practically converge to the optimal solution and capture the invariant features is that the optimizer is allowed to take advantage of the momentum of the gradients. In other words, momentum allows the optimizer to continue updating a parameter in the direction that might have been indicated several iterations ago, even if the gradients flowing in that parameter are zeroed out by AND-mask. More formally, if we model the optimizer as a function $g(.)$, the update received by component $j$ in the ILC work is 
\begin{eqnarray}
\begin{cases}
g\left(\tilde{\nabla\mathcal{L}_j^k}, M_j^k \right), &\text{if}~~ \Big|\sum_{e}^{}\text{sign}\big([\nabla\mathcal{L}_e]_j\big)\Big| \geq \tau \\
g\left(0, M_j^k\right), & \text{otherwise}
\end{cases}
\end{eqnarray}
where $\tilde{\nabla\mathcal{L}_j}$ is the mean of gradients from different environments, either arithmetic or geometric, and $M_j^k$ is the momentum of gradient at iteration $k$.  

\begin{figure}[t]
    \centering
    \subfloat[][]{\includegraphics[width=0.33\linewidth]{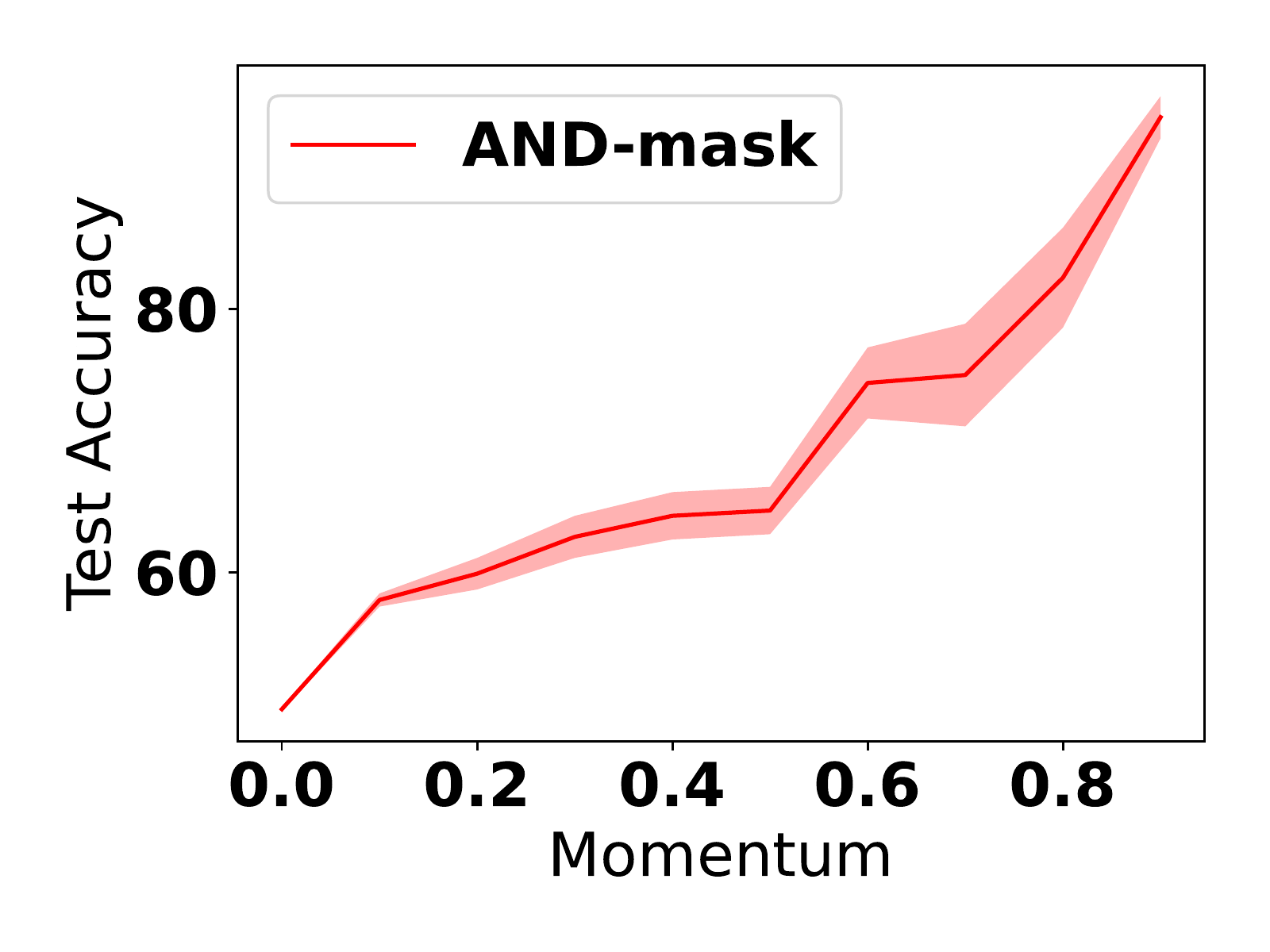}}
    \subfloat[][]{\includegraphics[width=0.33\linewidth]{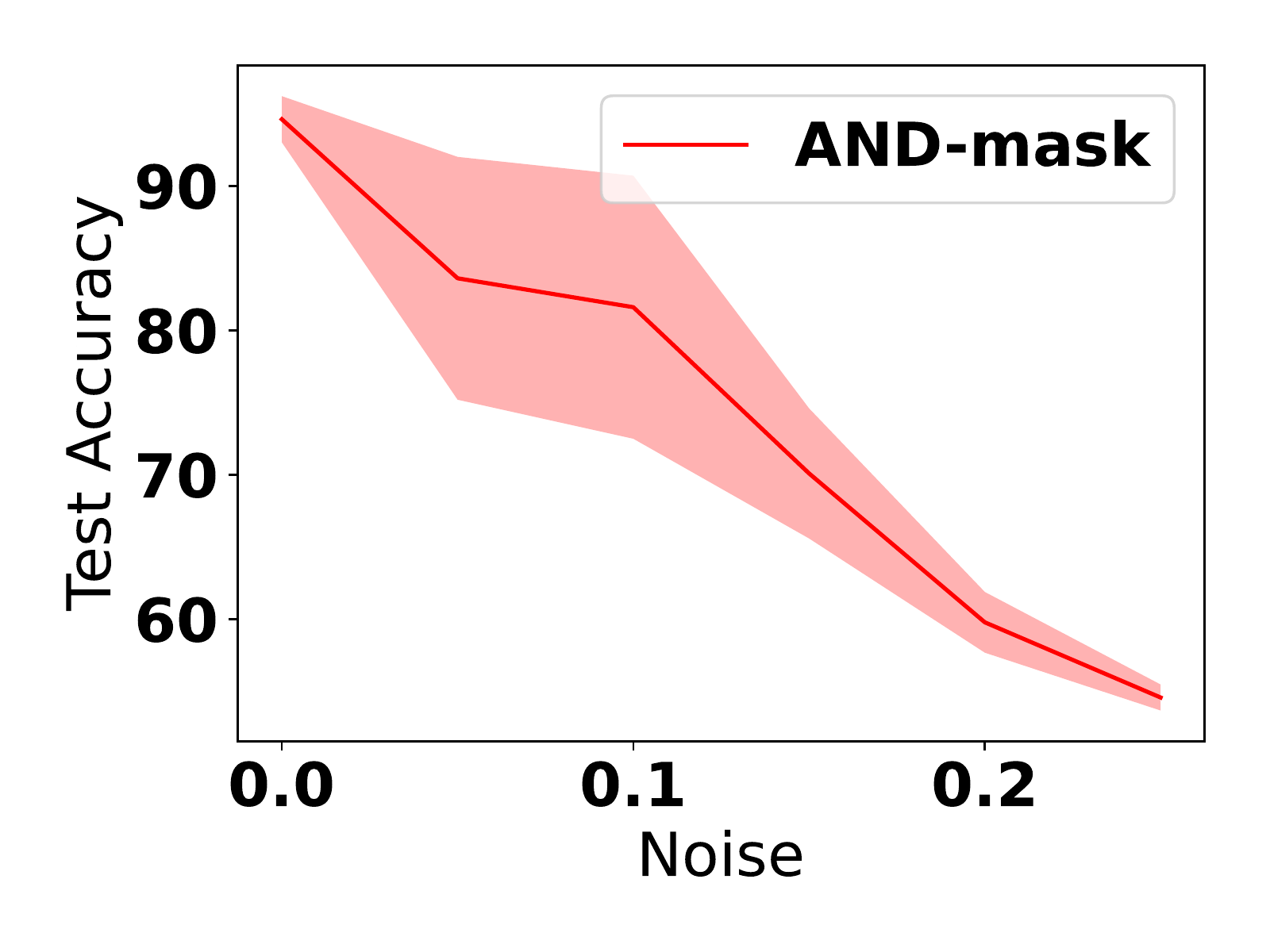}}
    \subfloat[][]{\includegraphics[width=0.33\linewidth]{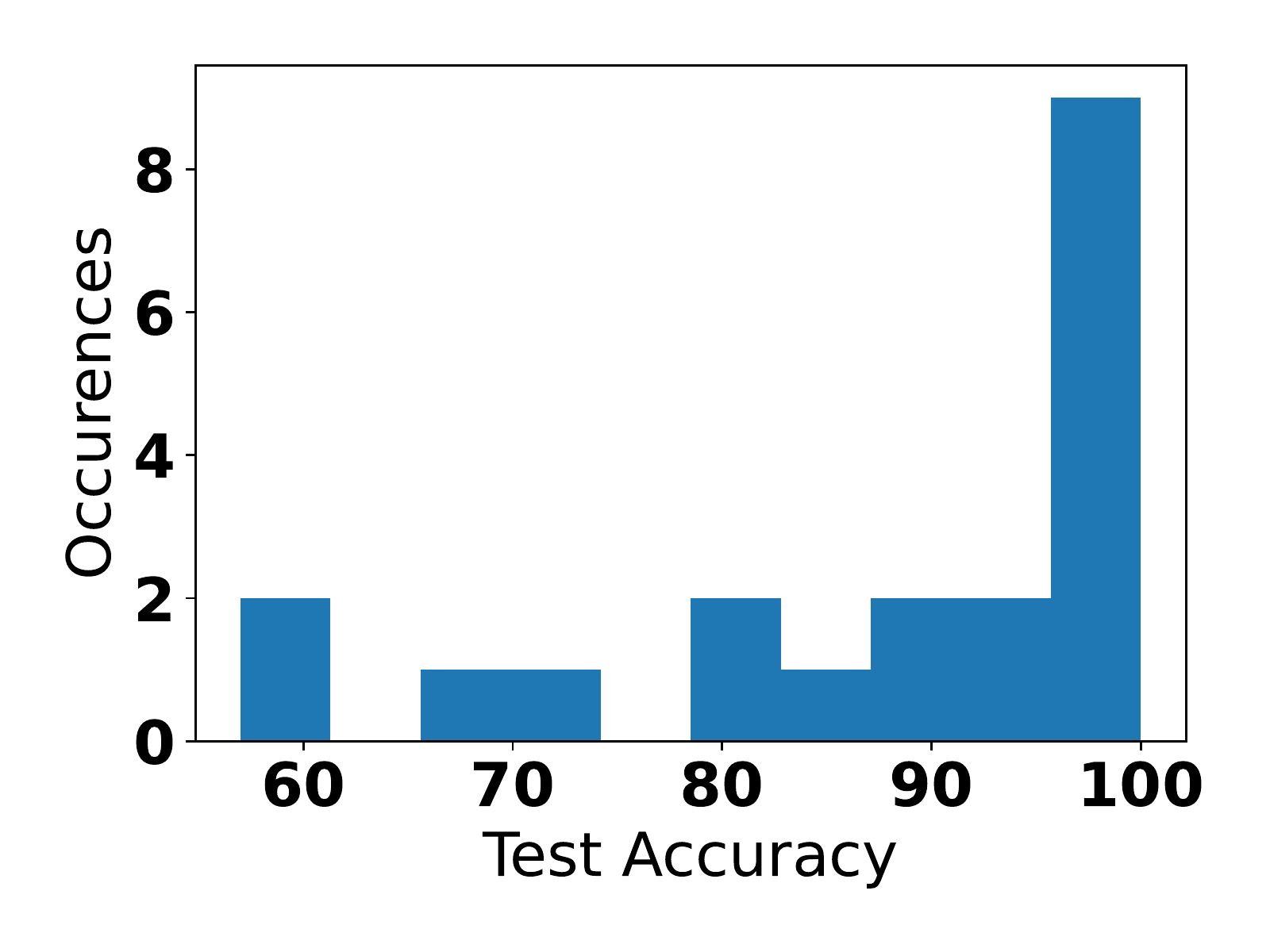}}
    \caption{Test accuracy of AND-mask on the Spirals dataset~\cite{ilc} over 20 seeds with best known hyperparameters of AND-mask on Spirals. (a) under different values of momentum; (b) under different amplitude of noise added to the invariant dimensions of the spirals dataset; (c) under different initializations. }
    \label{fig:momentum_fail}
\end{figure}

To verify this observation, the AND-masking is evaluated on the Spirals dataset~\cite{ilc} with varying values of momentum (see Fig.~\ref{fig:momentum_fail}~(a)). The fact that performance is so much correlated with the momentum tells us that the deadzones are present enough in the loss landscape as to considerably impact performance when no strategy is employed to circumvent them. One could say that the high dimensionality of the model parameter space should suffice as to avoid this problem, but the failure to solve the dataset with no momentum that it doesn't suffice as a valide escape strategy of the deadzones.

\paragraph{Sensitivity of AND-masking:} Employing a strict sign function to capture the direction of gradients followed by a heaviside function to form the AND-mask results in a sensitive process that can be deluded by infinitesimal gradients that might act as noise on the masking process. In the ILC regime, all the gradients, regardless of their magnitude, have equal contribution to the formation of the mask. This means that a small perturbation on an uncertain gradient of small magnitude can flip its sign and intrigue the AND-mask to zero out the whole set of gradients from all environment. \textcolor{black}{As the results in Fig.~\ref{fig:momentum_fail}~(b) suggest}, the AND-masking technique is highly sensitive to noisy data and a decrease in its performance is observed over different levels of noise.

\paragraph{Convergence is highly dependent on initialization:} The presence of dead zones and the fact that getting stuck in a dead zone is proportional to the value of $\tau$, render a high degree of importance for the initialization of the model. Initializing a model such that some of its parameters are already in a dead zone or are likely to get stuck in one significantly restricts the possible pool of solutions to which the model has access. This creates outlier solutions that are undesirable in the OOD scenario where we don't have access to the test set to know how our model generalizes to unseen trials. A preferable behavior would be to have less outliers to improve robustness on real-world datasets. To verify this, please refer to Fig.~\ref{fig:momentum_fail}~(c), where we have experimentally validated the susceptibility of AND-masking to initialization.



\subsection{The Proposed SAND-mask}

In this section, we propose an enhanced version of the AND-masking technique, referred to as ``SAND-mask'', which not only inherits the core idea behind the AND-masking technique, but also addresses its failure modes. In other words, as opposed to AND-mask that a strict criteria on matching the direction of gradients from all environments is applied, SAND-mask employs a smooth function capture and promote the invariant features among training domains. SAND-mask is formulated as
\begin{eqnarray}
m_{\tau} = \text{max} \left( 0, \text{tanh}\left(\frac{1}{\sigma^2}\left(\left|\frac{1}{|\mathcal{E}|}\sum_{e \in \mathcal{E}}^{} \text{sign}(\nabla\mathcal{L}_e)\right| - \tau \right)\right) \right),\label{eq:smoothilc}
\end{eqnarray} 
where $\tau$ is the agreement threshold the determines the fraction of environments that need to agree in terms of direction of their gradients. In addition, SAND-mask introduces a new parameter, $\sigma^2$, to measure the dispersion of the magnitude of gradients and encourages the agreement of magnitude as well as direction among the environments. $\sigma^2_j$ for each component of network and across all environments can be calculated as 
\begin{eqnarray}
\sigma^2_j = \frac{\text{var}(\nabla\mathcal{L}_j)}{\text{avg}(\nabla\mathcal{L}_j)^2}. 
\end{eqnarray}

\begin{SCfigure}
\includegraphics[width=0.4\linewidth]{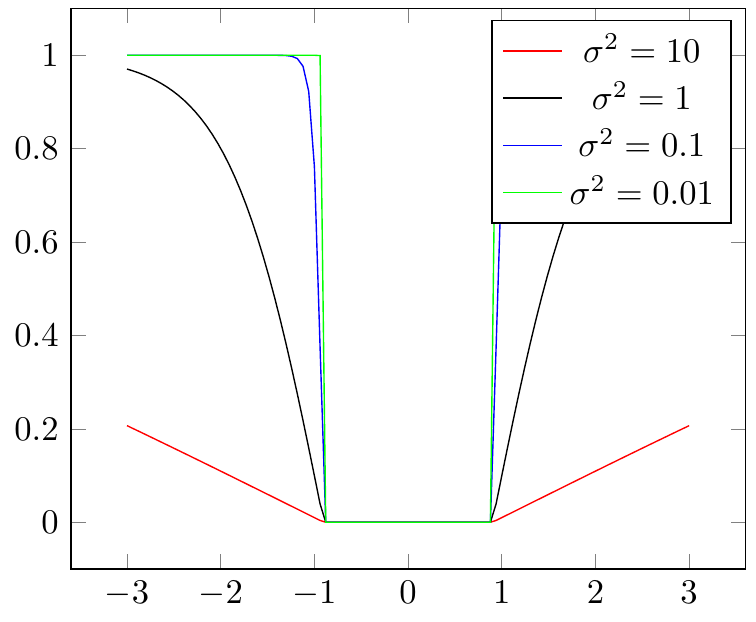}
\caption{Behavior of SAND-mask for different levels of agreement among the magnitude of gradients, when $\tau=0.9$. As it is understood, SAND-mask replicates the AND-mask only when there is a high level of agreement among the magnitude of the gradients. Otherwise, it lowers the weight assigned to the gradients to control the speed of training and avoid getting overconfident on spurious correlations in the training data.}\label{fig:smoothilc}
\end{SCfigure}

As shown in Fig.~\ref{fig:smoothilc}, SAND-mask changes its shape dynamically and based on the agreement among the magnitude of gradients such that a higher agreement is more likely to be weighted by ``$1$'' and vice versa. In other words, SAND-mask introduces a continuous weighting scheme to construct the mask that is in contrast with the Boolean mask created by the AND-masking technique. Since in SAND-mask the direction and the magnitude of gradients are simultaneously checked to verify the agreement among environments, one can reduce the probability of getting stuck in deadzones by lowering the agreement threshold, $\tau$. Please note that although lowering the agreement threshold might not be a reasonable strategy as we are interested in full agreement among environments but the dynamic behavior of SAND-mask that assigns a weight proportional to the degree of agreement among the magnitude of gradients would still help the model to capture the invariant features. As a final note, our proposed masking strategy significantly reduces the number of dead zones in the loss space by introducing a transient space between ``full-agreement'' and ``no-agreement'' cases (orthants) among the gradients. 


We believe that although the alignment of directions among gradients is a desirable/informative measure of consistency among environments, this property can be easily counterfeited by small/noisy/outlier gradients. In such cases, the consistency of magnitude could serve as an additional clue to decide if the gradients from different environments are confidently reporting a certain direction and update to the parameter or not.  


\section{Experiments}
In this section, the evaluation results of our proposed SAND-masking technique over well-known and popular domain generalization datasets is provided. In what follows, the benchmark on which our work is implemented and compared with other state-of-the-art techniques is introduced. 

\subsection{Domain Generalization Benchmark; DomainBed}
DomainBed~\cite{domainbed} provides a platform to study domain generalization of any algorithm across several benchmarking datasets and under a rigorous model selection and hyperparameter search. The core idea behind DomainBed is that the performance of domain generalization algorithms is heavily dependent on the architecture and the hyperparameters used. Therefore, DomainBed proposes three different model selection schemes based on how the validation set upon which the hyperparameters are fine tuned is formed. The three schemes are as follows:

\begin{itemize}

\item \textbf{Training-domain Validation Set:} In this scenario, the validation set is formed by randomly collecting $20\%$ of data from each of the training domains. This scenario imposes the most strict condition on hyperparameter fine tuning of the models.  

\item \textbf{Leave-one-domain-out Cross-validation:} Assuming that $n_e$ training domains are in hand, DomainBed trains $n_e$ models based on the training data from $n_e-1$ domains, while keeping the hyperparameters the same for all of the domains. Afterwards, the validation accuracy is reported as the average accuracy of all of the $n_e$ domains, which then will be the basis for fine tuning the hyperparameters. 

\item \textbf{Test-domain validation set (Oracle)} In this scenario, the validation set is formed based on the data in the test domains, and the hyper parameters are tuned based on test-time performance. However, to avoid rendering the problem as domain adaptation instead of domain generalization, access to the validation set is only feasible at the end of the training and therefore, early stopping of the training is not feasible. Please note that in this scenario, all the models based on different algorithms should undergo a fixed number of training steps to be fairly compared with each other. 
\end{itemize}

In this work, we have evaluated the SAND-mask based on the first and the third validation scenarios as the Leave-one-domain-out Cross-validation is considerably expensive from computational point of view. The datasets used for our evaluations include Colored MNIST~\cite{irm}, Rotated MNIST~\cite{rmnist}, VLCS~\cite{fang2013unbiased}, PACS~\cite{li2017deeper}, Terra Incognita~\cite{beery2018recognition} and Office-Home~\cite{venkateswara2017deep}. The summary of our evaluations of SAND-mask and comparisons with its counterparts are provided in Tables~\ref{table:1} and~\ref{table:2}, where the former summarizes the performance in ``training-domain validation set'', while the latter shows the performance in ``test-domain validation set''. The details of our experiments including the architectures employed and the range of hyperparameters for each model are given in Appendix~\ref{appendix:experiment}, and a comprehensive summary of the performances for all algorithms across all of the datasets is provided in Appendix~\ref{appendix:SupplementaryResults}.  

In addition to the above experiments, we have evaluated our work over the Spirals dataset~\cite{ilc}, which is the original testbed for the AND-masking technique. Due to the complex nature of this dataset and its considerably higher number of domains ($16$ environments) compared to other datasets, we have only compared SAND-mask with AND-mask, IRM, and the ERM technique, as the results can be found in Table~\ref{table:spiral}.  

\begin{table}
\caption{Model selection: Training-domain Validation Set}\label{table:1}
\adjustbox{max width=\textwidth}{%
\begin{tabular}{lcccccccc}
\toprule
\textbf{Algorithm}        & \textbf{ColoredMNIST}     & \textbf{RotatedMNIST}     & \textbf{VLCS}             & \textbf{PACS}             & \textbf{OfficeHome}       & \textbf{TerraIncognita}   & \textbf{DomainNet}        & \textbf{Avg}              \\
\midrule
ERM                       & 51.5 $\pm$ 0.1            & 98.0 $\pm$ 0.0            & 77.5 $\pm$ 0.4            & 85.5 $\pm$ 0.2            & 66.5 $\pm$ 0.3            & 46.1 $\pm$ 1.8            & 40.9 $\pm$ 0.1            & 66.6                      \\
IRM                       & 52.0 $\pm$ 0.1            & 97.7 $\pm$ 0.1            & 78.5 $\pm$ 0.5            & 83.5 $\pm$ 0.8            & 64.3 $\pm$ 2.2            & 47.6 $\pm$ 0.8            & 33.9 $\pm$ 2.8            & 65.4                      \\
GroupDRO                  & 52.1 $\pm$ 0.0            & 98.0 $\pm$ 0.0            & 76.7 $\pm$ 0.6            & 84.4 $\pm$ 0.8            & 66.0 $\pm$ 0.7            & 43.2 $\pm$ 1.1            & 33.3 $\pm$ 0.2            & 64.8                      \\
Mixup                     & 52.1 $\pm$ 0.2            & 98.0 $\pm$ 0.1            & 77.4 $\pm$ 0.6            & 84.6 $\pm$ 0.6            & 68.1 $\pm$ 0.3            & 47.9 $\pm$ 0.8            & 39.2 $\pm$ 0.1            & 66.7                      \\
MLDG                      & 51.5 $\pm$ 0.1            & 97.9 $\pm$ 0.0            & 77.2 $\pm$ 0.4            & 84.9 $\pm$ 1.0            & 66.8 $\pm$ 0.6            & 47.7 $\pm$ 0.9            & 41.2 $\pm$ 0.1            & 66.7                      \\
CORAL                     & 51.5 $\pm$ 0.1            & 98.0 $\pm$ 0.1            & 78.8 $\pm$ 0.6            & 86.2 $\pm$ 0.3            & 68.7 $\pm$ 0.3            & 47.6 $\pm$ 1.0            & 41.5 $\pm$ 0.1            & 67.5                      \\
MMD                       & 51.5 $\pm$ 0.2            & 97.9 $\pm$ 0.0            & 77.5 $\pm$ 0.9            & 84.6 $\pm$ 0.5            & 66.3 $\pm$ 0.1            & 42.2 $\pm$ 1.6            & 23.4 $\pm$ 9.5            & 63.3                      \\
DANN                      & 51.5 $\pm$ 0.3            & 97.8 $\pm$ 0.1            & 78.6 $\pm$ 0.4            & 83.6 $\pm$ 0.4            & 65.9 $\pm$ 0.6            & 46.7 $\pm$ 0.5            & 38.3 $\pm$ 0.1            & 66.1                      \\
CDANN                     & 51.7 $\pm$ 0.1            & 97.9 $\pm$ 0.1            & 77.5 $\pm$ 0.1            & 82.6 $\pm$ 0.9            & 65.8 $\pm$ 1.3            & 45.8 $\pm$ 1.6            & 38.3 $\pm$ 0.3            & 65.6                      \\
MTL                       & 51.4 $\pm$ 0.1            & 97.9 $\pm$ 0.0            & 77.2 $\pm$ 0.4            & 84.6 $\pm$ 0.5            & 66.4 $\pm$ 0.5            & 45.6 $\pm$ 1.2            & 40.6 $\pm$ 0.1            & 66.2                      \\
SagNet                    & 51.7 $\pm$ 0.0            & 98.0 $\pm$ 0.0            & 77.8 $\pm$ 0.5            & 86.3 $\pm$ 0.2            & 68.1 $\pm$ 0.1            & 48.6 $\pm$ 1.0            & 40.3 $\pm$ 0.1            & 67.2                      \\
ARM                       & 56.2 $\pm$ 0.2            & 98.2 $\pm$ 0.1            & 77.6 $\pm$ 0.3            & 85.1 $\pm$ 0.4            & 64.8 $\pm$ 0.3            & 45.5 $\pm$ 0.3            & 35.5 $\pm$ 0.2            & 66.1                      \\
VREx                      & 51.8 $\pm$ 0.1            & 97.9 $\pm$ 0.1            & 78.3 $\pm$ 0.2            & 84.9 $\pm$ 0.6            & 66.4 $\pm$ 0.6            & 46.4 $\pm$ 0.6            & 33.6 $\pm$ 2.9            & 65.6                      \\
RSC                       & 51.7 $\pm$ 0.2            & 97.6 $\pm$ 0.1            & 77.1 $\pm$ 0.5            & 85.2 $\pm$ 0.9            & 65.5 $\pm$ 0.9            & 46.6 $\pm$ 1.0            & 38.9 $\pm$ 0.5            & 66.1                      \\
\midrule
AND-mask                   & 51.3 $\pm$ 0.2       & 97.6 $\pm$ 0.1 & 78.1 $\pm$ 0.9 &84.4 $\pm$ 0.9& 65.6 $\pm$ 0.4 & 44.6 $\pm$ 0.3     & 37.2$\pm$ 0.6     & 65.5                      \\
SAND-mask                 & 51.8 $\pm$ 0.2       & 97.4 $\pm$ 0.1 & 77.4 $\pm$ 0.2 &84.6 $\pm$ 0.9& 65.8 $\pm$ 0.4&  42.9 $\pm$ 1.7  & 32.1$\pm$ 0.6        & 64.6                      \\
\bottomrule
\end{tabular}}
\end{table}

\begin{table}
\caption{Model Selection: Test-domain Validation Set (Oracle)}\label{table:2}
\adjustbox{max width=\textwidth}{%
\begin{tabular}{lcccccccc}
\toprule
\textbf{Algorithm}        & \textbf{ColoredMNIST}     & \textbf{RotatedMNIST}     & \textbf{VLCS}             & \textbf{PACS}             & \textbf{OfficeHome}       & \textbf{TerraIncognita}   & \textbf{DomainNet}        & \textbf{Avg}              \\
\midrule
ERM                       & 57.8 $\pm$ 0.2            & 97.8 $\pm$ 0.1            & 77.6 $\pm$ 0.3            & 86.7 $\pm$ 0.3            & 66.4 $\pm$ 0.5            & 53.0 $\pm$ 0.3            & 41.3 $\pm$ 0.1            & 68.7                      \\
IRM                       & 67.7 $\pm$ 1.2            & 97.5 $\pm$ 0.2            & 76.9 $\pm$ 0.6            & 84.5 $\pm$ 1.1            & 63.0 $\pm$ 2.7            & 50.5 $\pm$ 0.7            & 28.0 $\pm$ 5.1            & 66.9                      \\
GroupDRO                  & 61.1 $\pm$ 0.9            & 97.9 $\pm$ 0.1            & 77.4 $\pm$ 0.5            & 87.1 $\pm$ 0.1            & 66.2 $\pm$ 0.6            & 52.4 $\pm$ 0.1            & 33.4 $\pm$ 0.3            & 67.9                      \\
Mixup                     & 58.4 $\pm$ 0.2            & 98.0 $\pm$ 0.1            & 78.1 $\pm$ 0.3            & 86.8 $\pm$ 0.3            & 68.0 $\pm$ 0.2            & 54.4 $\pm$ 0.3            & 39.6 $\pm$ 0.1            & 69.0                      \\
MLDG                      & 58.2 $\pm$ 0.4            & 97.8 $\pm$ 0.1            & 77.5 $\pm$ 0.1            & 86.8 $\pm$ 0.4            & 66.6 $\pm$ 0.3            & 52.0 $\pm$ 0.1            & 41.6 $\pm$ 0.1            & 68.7                      \\
CORAL                     & 58.6 $\pm$ 0.5            & 98.0 $\pm$ 0.0            & 77.7 $\pm$ 0.2            & 87.1 $\pm$ 0.5            & 68.4 $\pm$ 0.2            & 52.8 $\pm$ 0.2            & 41.8 $\pm$ 0.1            & 69.2                      \\
MMD                       & 63.3 $\pm$ 1.3            & 98.0 $\pm$ 0.1            & 77.9 $\pm$ 0.1            & 87.2 $\pm$ 0.1            & 66.2 $\pm$ 0.3            & 52.0 $\pm$ 0.4            & 23.5 $\pm$ 9.4            & 66.9                      \\
DANN                      & 57.0 $\pm$ 1.0            & 97.9 $\pm$ 0.1            & 79.7 $\pm$ 0.5            & 85.2 $\pm$ 0.2            & 65.3 $\pm$ 0.8            & 50.6 $\pm$ 0.4            & 38.3 $\pm$ 0.1            & 67.7                      \\
CDANN                     & 59.5 $\pm$ 2.0            & 97.9 $\pm$ 0.0            & 79.9 $\pm$ 0.2            & 85.8 $\pm$ 0.8            & 65.3 $\pm$ 0.5            & 50.8 $\pm$ 0.6            & 38.5 $\pm$ 0.2            & 68.2                      \\
MTL                       & 57.6 $\pm$ 0.3            & 97.9 $\pm$ 0.1            & 77.7 $\pm$ 0.5            & 86.7 $\pm$ 0.2            & 66.5 $\pm$ 0.4            & 52.2 $\pm$ 0.4            & 40.8 $\pm$ 0.1            & 68.5                      \\
SagNet                    & 58.2 $\pm$ 0.3            & 97.9 $\pm$ 0.0            & 77.6 $\pm$ 0.1            & 86.4 $\pm$ 0.4            & 67.5 $\pm$ 0.2            & 52.5 $\pm$ 0.4            & 40.8 $\pm$ 0.2            & 68.7                      \\
ARM                       & 63.2 $\pm$ 0.7            & 98.1 $\pm$ 0.1            & 77.8 $\pm$ 0.3            & 85.8 $\pm$ 0.2            & 64.8 $\pm$ 0.4            & 51.2 $\pm$ 0.5            & 36.0 $\pm$ 0.2            & 68.1                      \\
VREx                      & 67.0 $\pm$ 1.3            & 97.9 $\pm$ 0.1            & 78.1 $\pm$ 0.2            & 87.2 $\pm$ 0.6            & 65.7 $\pm$ 0.3            & 51.4 $\pm$ 0.5            & 30.1 $\pm$ 3.7            & 68.2                      \\
RSC                       & 58.5 $\pm$ 0.5            & 97.6 $\pm$ 0.1            & 77.8 $\pm$ 0.6            & 86.2 $\pm$ 0.5            & 66.5 $\pm$ 0.6            & 52.1 $\pm$ 0.2            & 38.9 $\pm$ 0.6            & 68.2                      \\
\midrule
AND-mask                   & 58.6 $\pm$ 0.4     & 97.5 $\pm$ 0.0 & 76.4 $\pm$ 0.4 & 86.4 $\pm$ 0.4 & 66.1 $\pm$ 0.2& 49.8 $\pm$ 0.4      &37.9$\pm$ 0.6      & 67.5                      \\
SAND-mask                 & 62.3 $\pm$ 1.0     & 97.4 $\pm$ 0.1 & 76.2 $\pm$ 0.5 & 85.9 $\pm$ 0.4 &  65.9 $\pm$ 0.5 &    50.2 $\pm$ 0.1    & 32.3$\pm$ 0.6.      & 67.2                    \\

\bottomrule
\end{tabular}}
\end{table}

\begin{table}
\centering
\caption{Experiments on the Spiral dataset}~\label{table:spiral}
\adjustbox{max width=\textwidth}{%
\begin{tabular}{lcc}
\toprule
\textbf{Algorithm}        &\textbf{Training-domain Validation Set} &\textbf{Test-domain Validation Set (Oracle)} \\
\midrule
ERM                       & 45.8 $\pm$ 2.4            &   94.2 $\pm$ 1.3       \\
IRM                       & 54.7 $\pm$ 3.6            &   89.0 $\pm$ 1.2       \\
AND-mask                   & 88.0 $\pm$ 2.9            &   97.25 $\pm$ 0.3       \\
SAND-mask                 & 49.2 $\pm$ 5.4            &   91.1 $\pm$ 2.4       \\
\bottomrule
\end{tabular}}
\end{table}

\section{Discussion}
In this work, we introduced Smoothed-AND (SAND)-masking technique that improves the performance of the current state-of-the-art OOD methods over a variety of datasets. In fact, SAND-mask aims at addressing the failure modes that we identified for a recent major contributions in the field of OOD generalization, i.e., Reference~\cite{ilc}. As it is supported by a rigorous and exhaustive set of results on the DomainBed benchmark, SAND-mask outperforms its counterparts and significantly enhances the classification accuracy over the Colored MNIST dataset for about $6\%$. Despite the superior performance of SAND-mask over different datasets, in what follows, we elaborate on its limitations and potential direction to be pursued in future. 

\noindent
\paragraph{Limitations and Future Work:}
\label{limitation_of_work}(\textbf{1}) Although SAND-mask aims at replicating the behavior of AND-masking technique, the results in Table~\ref{table:spiral} show that it barely matches the performance of AND-mask over the Spiral dataset. However and on the other hand, SAND-mask manages to outperform or perform similarly to AND-mask on other OOD generalization datasets, especially on the Colored MNIST dataset that there is a performance gap of $~15\%$. This behavior needs to be further investigated as it seems that Spiral dataset looks at OOD generalization from a different point of view than other datasets in the field. Our SAND-mask technique has shown the capacity to bridge the two views and provide a more general solution but it needs to be studied in more depth.
(\textbf{2}) Since the masking strategies, including AND-mask and SAND-mask, are not a penalty to the objective function, upon their satisfaction in the training phase, they are unable of stopping the training procedure as the objective function might still calculate a considerable loss over the training data. Therefore, this property leads to a self destructive behavior during training as the objective function is always trying to pull the model towards a lower loss, even though it is not optimal for generalization of the model. This limitation also requires further attention so that the objective function can get a signal of how matched the Hessians of the environments have became during the training. 

\paragraph{Broader Impacts:}\label{Broader_impact} The collective efforts towards enhancing the performance of machine learning models in OOD settings is to further facilitate their widespread application in different domains, without the need to collect new domain-specific datasets and retraining the model. For instance, OOD generalization techniques can improve the performance of the deep learning models that are used in medical settings, e.g. medical imaging, as the differences in the skin tone, body shape and so many other factors related to people and diseases across the globe resemble the out of domain/distribution setting that we investigate in our experiments. In addition, the power of OOD methods in disentangling the invariant features from the ones that are spuriously correlated with the data can significantly help reducing the medical errors that happen when the training data for such models is not homogeneously collected. 


Another positive impact of considering OOD methods, in general, is that facial recognition systems are prone to shift from training distribution, mainly when the source of these shifts is related to skin tone or color. So all efforts towards reducing the failure of existing OOD approaches can overcome the possible discrimination caused by facial recognition systems that are vastly used in different applications. 

\newpage
\small
\bibliographystyle{unsrtnat}
\bibliography{main-v1.bib}

\newpage
\appendix

\section{Experiment}\label{appendix:experiment}
\subsection{Implementation Details}
Experiment where performed on the DomainBed~\cite{domainbed} suite\footnote{https://github.com/facebookresearch/DomainBed} (MIT License) with the added Spirals dataset. All experimentation where done on an internal cluster over 1 week on 50 NVIDIA Quadro RTX 8000 GPUs.
\paragraph{Hyperparameter Search} For each algorithm and test environment for a given dataset we perform a random search of hyperparameters over 20 sampled configurations from distributions (see Table~\ref{table:hyperparameters}). We split the data from each domain into 80\% and 20\% splits. We use the 80\% split for training and final evaluation of the model and use the hidden 20\% split for hyperparameter selection process.

\begin{table}[h]
    \caption{Hyperparameters, their default values and distributions for random search for the AND-mask and SAND-mask algorithms. Hyperparameter search space of baselines can be found in the original work of \citet{domainbed} from which results were taken from.} 
    \begin{center}
    { 
    \begin{tabular}{llll}
        \toprule
        \textbf{Condition} & \textbf{Parameter} & \textbf{Default value} & \textbf{Random distribution}\\
        \midrule
        VLCS / PACS                 & learning rate & 0.00005 & $10^{\text{Uniform}(-5, -3.5)}$\\
        TerraIncognita              & batch size    & 32   & $2^{\text{Uniform}(3, 5.5)}$\\
        OfficeHome                  & weight decay & 0    & $10^{\text{Uniform}(-6, -2)}$\\
        \midrule
        
        Rotated MNIST       & learning rate & 0.001 & $10^{\text{Uniform}(-4.5, -3.5)}$\\
        Colored MNIST       & batch size    & 64   & $2^{\text{Uniform}(3, 9)}$\\
                            & weight decay & 0    & $0$\\
        
        \midrule
        
        \multirow{5}{*}{Spirals}         & learning rate & 0.01 & $10^{\text{Uniform}(-3.5, -1.5)}$\\
                        & batch size    & 512   & $2^{\text{Uniform}(3, 9)}$\\
                        & weight decay & $0.001$    & $10^{\text{Uniform}(-6, -2)}$\\
                        & MLP depth & 3 & $\text{RandomChoice}([3, 4, 5])$\\
                        & MLP width & 256 & $2^{\text{Uniform}([6,10])}$\\
                        
        \midrule
        AND-mask                &\multirow{2}{*}{ $\tau$ }& \multirow{2}{*}{1} & \multirow{2}{*}{$\text{Uniform}(0, 1)$}\\
        SAND-mask                \\
        \midrule
        All          & dropout & 0    & $\text{RandomChoice}([0, 0.1, 0.5])$\\
        \bottomrule
    \end{tabular}
    }
    \end{center}
    \label{table:hyperparameters}
\end{table}
\paragraph{Error bars} For each of the 20 sampled hyperparameter configuration of algorithm and test environment pairs for a given dataset, we test 3 different seeds in order to standardize the performance of a given configuration and give us estimated error bars.
\paragraph{Baselines} Baselines for the datasets were taken directly from the most recent set of results of the DomainBed~\cite{domainbed} suite. Results for AND-mask and SAND-mask on all the datasets were obtained under the exact same setup which allows us to compare both set of results. Here is the list of algorithms that SAND-mask is compared with.

\begin{itemize}
\item \textbf{ERM: }Empirical Risk Minimization by~\citet{erm}
\item \textbf{IRM: }Invariant Risk Minimization by~\citet{irm}
\item \textbf{GroupDRO: }Group Distributionally Robust Optimization by~\citet{groupdro}
\item \textbf{Mixup: }Interdomain Mixup by~\citet{mixup}
\item \textbf{MTL: }Marginal Transfer Learning by~\citet{mtl}
\item \textbf{MLDG: }Meta Learning Domain Generalization by~\citet{mldg}
\item \textbf{MMD: }Maximum Mean Discrepancy by~\citet{mmd}
\item \textbf{CORAL: }Deep CORAL by~\citet{coral}
\item \textbf{DANN: }Domain Adversarial Neural Network by~\citet{dann}
\item \textbf{CDANN: }Conditional Domain Adversarial Neural Network by~\citet{cdann}
\item \textbf{SagNet: }Style Agnostic Networks by~\citet{sagnet}
\item \textbf{ARM: }Adaptive Risk Minimization by~\citet{arm}
\item \textbf{VREx: }Variance Risk Extrapolation by~\citet{rex}
\item \textbf{RSC: }Representation Self-Challenging by~\citet{rsc}
\item \textbf{SD: }Spectral Decoupling by~\citet{sd}
\item \textbf{AND-mask: }Learning Explanations that are Hard to Vary by~\citet{ilc}
\end{itemize}

\paragraph{Employed Architecture}
In Table~\ref{table:architectures}, we detail the architecture used for experimentation. For the MLP architecture, it's depth and width are defined as hyperparameter included in the hyperparameter search. For the ResNet-50 architecture, we use a ResNet-50 model pretrained on ImageNet of which we replace the final layer and fine-tune. The details regarding the architecture of the MNSIT ConvNet are given in Table~\ref{table:mnist_convnet}.
\begin{table}[h]
\centering
\begin{minipage}{0.45\textwidth}
    \caption{Neural network architectures used for each dataset.} 
        \begin{tabular}{ll}
        \toprule
        \textbf{Dataset} & \textbf{Architecture} \\
        \midrule
        Spirals & MLP \\
        \midrule
        Colored MNIST & \multirow{2}{*}{MNIST ConvNet} \\
        Rotated MNIST & \\
        \midrule
        PACS & \multirow{5}{*}{ResNet-50} \\
        VLCS & \\
        Office-Home &  \\
        TerraIncognita &  \\
        DomainNet &  \\
        \bottomrule
        \end{tabular}
    \label{table:architectures}
\end{minipage}
\hfill
\begin{minipage}{0.48\textwidth}
    \caption{Details of our MNIST ConvNet architecture. All convolutions use 3 $\times$ 3 kernels and ``same'' padding.} 
        \begin{tabular}{ll}
        \toprule
        \textbf{\#} & \textbf{Layer}\\
        \midrule
            1  & Conv2D (in=$d$, out=64)\\
            2  & ReLU\\
            3  & GroupNorm (groups=8)\\
            4  & Conv2D (in=64, out=128, stride=2)\\
            5  & ReLU\\
            6  & GroupNorm (8 groups)\\
            7  & Conv2D (in=128, out=128)\\
            8  & ReLU\\
            9  & GroupNorm (8 groups)\\
            10 & Conv2D (in=128, out=128)\\
            11 & ReLU\\
            12 & GroupNorm (8 groups)\\
            13 & Global average-pooling\\
        \bottomrule
        \end{tabular}
    \label{table:mnist_convnet}
\end{minipage}
\end{table}

\newpage
\section{Supplementary Results}\label{appendix:SupplementaryResults}

\subsection{Spirals}

\begin{center}
\adjustbox{max width=\linewidth}{%
\begin{tabular}{lccccccccccccccccc}
\toprule
\multicolumn{18}{c}{\textbf{Model selection method: training domain validation set}} \\
\midrule
\textbf{Algorithm}   & \textbf{0}           & \textbf{1}           & \textbf{2}           & \textbf{3}           & \textbf{4}           & \textbf{5}           & \textbf{6}           & \textbf{7}           & \textbf{8}           & \textbf{9}           & \textbf{10}          & \textbf{11}          & \textbf{12}          & \textbf{13}          & \textbf{14}          & \textbf{15}          & \textbf{Avg}         \\
\midrule
ERM                  & 51.4 $\pm$ 0.8       & 55.4 $\pm$ 18.4      & 45.8 $\pm$ 22.0      & 40.9 $\pm$ 4.9       & 20.9 $\pm$ 11.9      & 59.3 $\pm$ 6.8       & 36.5 $\pm$ 10.1      & 44.8 $\pm$ 15.3      & 33.3 $\pm$ 27.2      & 17.5 $\pm$ 14.0      & 68.0 $\pm$ 5.0       & 62.7 $\pm$ 15.3      & 54.5 $\pm$ 5.3       & 27.8 $\pm$ 16.1      & 49.1 $\pm$ 21.9      & 64.6 $\pm$ 26.4      & 45.8                 \\
IRM                  & 50.4 $\pm$ 0.9       & 37.5 $\pm$ 15.7      & 80.3 $\pm$ 11.5      & 77.7 $\pm$ 12.2      & 54.7 $\pm$ 23.3      & 64.8 $\pm$ 14.8      & 16.3 $\pm$ 12.2      & 49.6 $\pm$ 23.4      & 49.1 $\pm$ 20.3      & 45.1 $\pm$ 18.7      & 25.7 $\pm$ 10.4      & 48.3 $\pm$ 21.2      & 71.5 $\pm$ 12.0      & 58.2 $\pm$ 18.2      & 79.3 $\pm$ 12.3      & 66.1 $\pm$ 27.0      & 54.7                 \\
AND-mask              & 52.6 $\pm$ 0.5       & 99.7 $\pm$ 0.2       & 100.0 $\pm$ 0.0      & 99.6 $\pm$ 0.4       & 66.4 $\pm$ 27.1      & 99.9 $\pm$ 0.1       & 100.0 $\pm$ 0.0      & 100.0 $\pm$ 0.0      & 100.0 $\pm$ 0.0      & 100.0 $\pm$ 0.0      & 97.8 $\pm$ 1.0       & 99.7 $\pm$ 0.1       & 82.1 $\pm$ 14.6      & 45.4 $\pm$ 23.5      & 99.8 $\pm$ 0.2       & 66.4 $\pm$ 27.1      & 88.0                 \\
SAND-mask            & 51.9 $\pm$ 0.4       & 1.1 $\pm$ 0.9        & 33.3 $\pm$ 27.2      & 56.2 $\pm$ 18.3      & 33.3 $\pm$ 13.6      & 44.4 $\pm$ 22.9      & 74.9 $\pm$ 19.8      & 60.0 $\pm$ 17.8      & 37.2 $\pm$ 11.0      & 89.9 $\pm$ 8.3       & 69.0 $\pm$ 9.0       & 49.1 $\pm$ 2.1       & 66.6 $\pm$ 13.7      & 80.0 $\pm$ 16.3      & 25.5 $\pm$ 20.6      & 16.4 $\pm$ 13.4      & 49.2                 \\
\bottomrule
\end{tabular}}
\end{center}

\begin{center}
\adjustbox{max width=\linewidth}{%
\begin{tabular}{lccccccccccccccccc}
\toprule
\multicolumn{18}{c}{\textbf{Model selection method: test-domain validation set \textit{(oracle)}}} \\
\midrule
\textbf{Algorithm}   & \textbf{0}           & \textbf{1}           & \textbf{2}           & \textbf{3}           & \textbf{4}           & \textbf{5}           & \textbf{6}           & \textbf{7}           & \textbf{8}           & \textbf{9}           & \textbf{10}          & \textbf{11}          & \textbf{12}          & \textbf{13}          & \textbf{14}          & \textbf{15}          & \textbf{Avg}         \\
\midrule
ERM                  & 49.3 $\pm$ 0.3       & 98.3 $\pm$ 1.0       & 97.8 $\pm$ 1.8       & 100.0 $\pm$ 0.0      & 100.0 $\pm$ 0.0      & 91.6 $\pm$ 6.8       & 94.2 $\pm$ 4.7       & 100.0 $\pm$ 0.0      & 99.3 $\pm$ 0.5       & 96.5 $\pm$ 2.9       & 92.1 $\pm$ 6.4       & 95.8 $\pm$ 1.9       & 100.0 $\pm$ 0.0      & 99.0 $\pm$ 0.8       & 96.2 $\pm$ 2.3       & 97.4 $\pm$ 2.1       & 94.2                 \\
IRM                  & 50.1 $\pm$ 0.4       & 100.0 $\pm$ 0.0      & 83.1 $\pm$ 7.8       & 99.3 $\pm$ 0.6       & 98.7 $\pm$ 1.1       & 90.5 $\pm$ 7.8       & 92.6 $\pm$ 4.2       & 90.5 $\pm$ 7.7       & 93.0 $\pm$ 1.0       & 86.8 $\pm$ 10.8      & 100.0 $\pm$ 0.0      & 90.8 $\pm$ 7.5       & 87.5 $\pm$ 10.2      & 75.9 $\pm$ 5.5       & 98.2 $\pm$ 1.5       & 86.3 $\pm$ 2.5       & 89.0                 \\


AND-mask              & 63.5 $\pm$ 6.0       & 100.0 $\pm$ 0.0      & 100.0 $\pm$ 0.0      & 95.5 $\pm$ 3.7       & 100.0 $\pm$ 0.0      & 100.0 $\pm$ 0.0      & 100.0 $\pm$ 0.0      & 100.0 $\pm$ 0.0      & 100.0 $\pm$ 0.0      & 100.0 $\pm$ 0.0      & 100.0 $\pm$ 0.0      & 100.0 $\pm$ 0.0      & 100.0 $\pm$ 0.0      & 97.0 $\pm$ 2.4       & 100.0 $\pm$ 0.0      & 100.0 $\pm$ 0.0      & 97.25                 \\
SAND-mask            & 56.2 $\pm$ 1.6       & 83.3 $\pm$ 13.6      & 100.0 $\pm$ 0.0      & 100.0 $\pm$ 0.0      & 100.0 $\pm$ 0.0      & 100.0 $\pm$ 0.0      & 96.2 $\pm$ 3.1       & 98.5 $\pm$ 1.3       & 83.0 $\pm$ 13.8      & 91.9 $\pm$ 6.4       & 100.0 $\pm$ 0.0      & 83.0 $\pm$ 13.9      & 84.9 $\pm$ 6.7       & 100.0 $\pm$ 0.0      & 85.6 $\pm$ 11.8      & 100.0 $\pm$ 0.0      & 91.1                 \\
\bottomrule
\end{tabular}}
\end{center}

\subsection{ColoredMNIST}

\begin{center}
\adjustbox{max width=\textwidth}{%
\begin{tabular}{lcccc}
\toprule
\multicolumn{5}{c}{\textbf{Model selection method: training domain validation set}} \\
\midrule
\textbf{Algorithm}   & \textbf{+90\%}       & \textbf{+80\%}       & \textbf{-90\%}       & \textbf{Avg}         \\
\midrule
ERM                  & 71.7 $\pm$ 0.1       & 72.9 $\pm$ 0.2       & 10.0 $\pm$ 0.1       & 51.5                 \\
IRM                  & 72.5 $\pm$ 0.1       & 73.3 $\pm$ 0.5       & 10.2 $\pm$ 0.3       & 52.0                 \\
GroupDRO             & 73.1 $\pm$ 0.3       & 73.2 $\pm$ 0.2       & 10.0 $\pm$ 0.2       & 52.1                 \\
Mixup                & 72.7 $\pm$ 0.4       & 73.4 $\pm$ 0.1       & 10.1 $\pm$ 0.1       & 52.1                 \\
MLDG                 & 71.5 $\pm$ 0.2       & 73.1 $\pm$ 0.2       & 9.8 $\pm$ 0.1        & 51.5                 \\
CORAL                & 71.6 $\pm$ 0.3       & 73.1 $\pm$ 0.1       & 9.9 $\pm$ 0.1        & 51.5                 \\
MMD                  & 71.4 $\pm$ 0.3       & 73.1 $\pm$ 0.2       & 9.9 $\pm$ 0.3        & 51.5                 \\
DANN                 & 71.4 $\pm$ 0.9       & 73.1 $\pm$ 0.1       & 10.0 $\pm$ 0.0       & 51.5                 \\
CDANN                & 72.0 $\pm$ 0.2       & 73.0 $\pm$ 0.2       & 10.2 $\pm$ 0.1       & 51.7                 \\
MTL                  & 70.9 $\pm$ 0.2       & 72.8 $\pm$ 0.3       & 10.5 $\pm$ 0.1       & 51.4                 \\
SagNet               & 71.8 $\pm$ 0.2       & 73.0 $\pm$ 0.2       & 10.3 $\pm$ 0.0       & 51.7                 \\
ARM                  & 82.0 $\pm$ 0.5       & 76.5 $\pm$ 0.3       & 10.2 $\pm$ 0.0       & 56.2                 \\
VREx                 & 72.4 $\pm$ 0.3       & 72.9 $\pm$ 0.4       & 10.2 $\pm$ 0.0       & 51.8                 \\
RSC                  & 71.9 $\pm$ 0.3       & 73.1 $\pm$ 0.2       & 10.0 $\pm$ 0.2       & 51.7                 \\
\midrule
AND-mask              & 70.7 $\pm$ 0.5       & 73.3 $\pm$ 0.2       & 10.0 $\pm$ 0.1       & 51.3                 \\
SAND-mask            & 72.0 $\pm$ 0.5       & 73.2 $\pm$ 0.4       & 10.3 $\pm$ 0.2       & 51.8                 \\
\bottomrule
\end{tabular}}
\end{center}

\begin{center}
\adjustbox{max width=\textwidth}{%
\begin{tabular}{lcccc}
\toprule
\multicolumn{5}{c}{\textbf{Model selection method: test-domain validation set \textit{(oracle)}}} \\
\midrule
\textbf{Algorithm}   & \textbf{+90\%}       & \textbf{+80\%}       & \textbf{-90\%}       & \textbf{Avg}         \\
\midrule
ERM                  & 71.8 $\pm$ 0.4       & 72.9 $\pm$ 0.1       & 28.7 $\pm$ 0.5       & 57.8                 \\
IRM                  & 72.0 $\pm$ 0.1       & 72.5 $\pm$ 0.3       & 58.5 $\pm$ 3.3       & 67.7                 \\
GroupDRO             & 73.5 $\pm$ 0.3       & 73.0 $\pm$ 0.3       & 36.8 $\pm$ 2.8       & 61.1                 \\
Mixup                & 72.5 $\pm$ 0.2       & 73.9 $\pm$ 0.4       & 28.6 $\pm$ 0.2       & 58.4                 \\
MLDG                 & 71.9 $\pm$ 0.3       & 73.5 $\pm$ 0.2       & 29.1 $\pm$ 0.9       & 58.2                 \\
CORAL                & 71.1 $\pm$ 0.2       & 73.4 $\pm$ 0.2       & 31.1 $\pm$ 1.6       & 58.6                 \\
MMD                  & 69.0 $\pm$ 2.3       & 70.4 $\pm$ 1.6       & 50.6 $\pm$ 0.2       & 63.3                 \\
DANN                 & 72.4 $\pm$ 0.5       & 73.9 $\pm$ 0.5       & 24.9 $\pm$ 2.7       & 57.0                 \\
CDANN                & 71.8 $\pm$ 0.5       & 72.9 $\pm$ 0.1       & 33.8 $\pm$ 6.4       & 59.5                 \\
MTL                  & 71.2 $\pm$ 0.2       & 73.5 $\pm$ 0.2       & 28.0 $\pm$ 0.6       & 57.6                 \\
SagNet               & 72.1 $\pm$ 0.3       & 73.2 $\pm$ 0.3       & 29.4 $\pm$ 0.5       & 58.2                 \\
ARM                  & 84.9 $\pm$ 0.9       & 76.8 $\pm$ 0.6       & 27.9 $\pm$ 2.1       & 63.2                 \\
VREx                 & 72.8 $\pm$ 0.3       & 73.0 $\pm$ 0.3       & 55.2 $\pm$ 4.0       & 67.0                 \\
RSC                  & 72.0 $\pm$ 0.1       & 73.2 $\pm$ 0.1       & 30.2 $\pm$ 1.6       & 58.5                 \\
\midrule
AND-mask              & 71.9 $\pm$ 0.6       & 73.6 $\pm$ 0.5       & 30.2 $\pm$ 1.4       & 58.6                 \\
SAND-mask            & 79.9 $\pm$ 3.8       & 75.9 $\pm$ 1.6       & 31.6 $\pm$ 1.1       & 62.3                 \\
\bottomrule
\end{tabular}}
\end{center}

\subsection{RotatedMNIST}

\begin{center}
\adjustbox{max width=\textwidth}{%
\begin{tabular}{lccccccc}
\toprule
\multicolumn{8}{c}{\textbf{Model selection method: training domain validation set}} \\
\midrule
\textbf{Algorithm}   & \textbf{0}           & \textbf{15}          & \textbf{30}          & \textbf{45}          & \textbf{60}          & \textbf{75}          & \textbf{Avg}         \\
\midrule
ERM                  & 95.9 $\pm$ 0.1       & 98.9 $\pm$ 0.0       & 98.8 $\pm$ 0.0       & 98.9 $\pm$ 0.0       & 98.9 $\pm$ 0.0       & 96.4 $\pm$ 0.0       & 98.0                 \\
IRM                  & 95.5 $\pm$ 0.1       & 98.8 $\pm$ 0.2       & 98.7 $\pm$ 0.1       & 98.6 $\pm$ 0.1       & 98.7 $\pm$ 0.0       & 95.9 $\pm$ 0.2       & 97.7                 \\
GroupDRO             & 95.6 $\pm$ 0.1       & 98.9 $\pm$ 0.1       & 98.9 $\pm$ 0.1       & 99.0 $\pm$ 0.0       & 98.9 $\pm$ 0.0       & 96.5 $\pm$ 0.2       & 98.0                 \\
Mixup                & 95.8 $\pm$ 0.3       & 98.9 $\pm$ 0.0       & 98.9 $\pm$ 0.0       & 98.9 $\pm$ 0.0       & 98.8 $\pm$ 0.1       & 96.5 $\pm$ 0.3       & 98.0                 \\
MLDG                 & 95.8 $\pm$ 0.1       & 98.9 $\pm$ 0.1       & 99.0 $\pm$ 0.0       & 98.9 $\pm$ 0.1       & 99.0 $\pm$ 0.0       & 95.8 $\pm$ 0.3       & 97.9                 \\
CORAL                & 95.8 $\pm$ 0.3       & 98.8 $\pm$ 0.0       & 98.9 $\pm$ 0.0       & 99.0 $\pm$ 0.0       & 98.9 $\pm$ 0.1       & 96.4 $\pm$ 0.2       & 98.0                 \\
MMD                  & 95.6 $\pm$ 0.1       & 98.9 $\pm$ 0.1       & 99.0 $\pm$ 0.0       & 99.0 $\pm$ 0.0       & 98.9 $\pm$ 0.0       & 96.0 $\pm$ 0.2       & 97.9                 \\
DANN                 & 95.0 $\pm$ 0.5       & 98.9 $\pm$ 0.1       & 99.0 $\pm$ 0.0       & 99.0 $\pm$ 0.1       & 98.9 $\pm$ 0.0       & 96.3 $\pm$ 0.2       & 97.8                 \\
CDANN                & 95.7 $\pm$ 0.2       & 98.8 $\pm$ 0.0       & 98.9 $\pm$ 0.1       & 98.9 $\pm$ 0.1       & 98.9 $\pm$ 0.1       & 96.1 $\pm$ 0.3       & 97.9                 \\
MTL                  & 95.6 $\pm$ 0.1       & 99.0 $\pm$ 0.1       & 99.0 $\pm$ 0.0       & 98.9 $\pm$ 0.1       & 99.0 $\pm$ 0.1       & 95.8 $\pm$ 0.2       & 97.9                 \\
SagNet               & 95.9 $\pm$ 0.3       & 98.9 $\pm$ 0.1       & 99.0 $\pm$ 0.1       & 99.1 $\pm$ 0.0       & 99.0 $\pm$ 0.1       & 96.3 $\pm$ 0.1       & 98.0                 \\
ARM                  & 96.7 $\pm$ 0.2       & 99.1 $\pm$ 0.0       & 99.0 $\pm$ 0.0       & 99.0 $\pm$ 0.1       & 99.1 $\pm$ 0.1       & 96.5 $\pm$ 0.4       & 98.2                 \\
VREx                 & 95.9 $\pm$ 0.2       & 99.0 $\pm$ 0.1       & 98.9 $\pm$ 0.1       & 98.9 $\pm$ 0.1       & 98.7 $\pm$ 0.1       & 96.2 $\pm$ 0.2       & 97.9                 \\
RSC                  & 94.8 $\pm$ 0.5       & 98.7 $\pm$ 0.1       & 98.8 $\pm$ 0.1       & 98.8 $\pm$ 0.0       & 98.9 $\pm$ 0.1       & 95.9 $\pm$ 0.2       & 97.6                 \\
\midrule
AND-mask              & 94.8 $\pm$ 0.2       & 98.8 $\pm$ 0.1       & 98.9 $\pm$ 0.0       & 98.7 $\pm$ 0.0       & 98.7 $\pm$ 0.1       & 95.5 $\pm$ 0.4       & 97.6                 \\
SAND-mask            & 94.5 $\pm$ 0.4       & 98.6 $\pm$ 0.1       & 98.8 $\pm$ 0.1       & 98.7 $\pm$ 0.1       & 98.6 $\pm$ 0.0       & 95.5 $\pm$ 0.2       & 97.4                 \\
\bottomrule
\end{tabular}}
\end{center}

\begin{center}
\adjustbox{max width=\textwidth}{%
\begin{tabular}{lccccccc}
\toprule
\multicolumn{8}{c}{\textbf{Model selection method: test-domain validation set \textit{(oracle)}}} \\
\midrule
\textbf{Algorithm}   & \textbf{0}           & \textbf{15}          & \textbf{30}          & \textbf{45}          & \textbf{60}          & \textbf{75}          & \textbf{Avg}         \\
\midrule
ERM                  & 95.3 $\pm$ 0.2       & 98.7 $\pm$ 0.1       & 98.9 $\pm$ 0.1       & 98.7 $\pm$ 0.2       & 98.9 $\pm$ 0.0       & 96.2 $\pm$ 0.2       & 97.8                 \\
IRM                  & 94.9 $\pm$ 0.6       & 98.7 $\pm$ 0.2       & 98.6 $\pm$ 0.1       & 98.6 $\pm$ 0.2       & 98.7 $\pm$ 0.1       & 95.2 $\pm$ 0.3       & 97.5                 \\
GroupDRO             & 95.9 $\pm$ 0.1       & 99.0 $\pm$ 0.1       & 98.9 $\pm$ 0.1       & 98.8 $\pm$ 0.1       & 98.6 $\pm$ 0.1       & 96.3 $\pm$ 0.4       & 97.9                 \\
Mixup                & 95.8 $\pm$ 0.3       & 98.7 $\pm$ 0.0       & 99.0 $\pm$ 0.1       & 98.8 $\pm$ 0.1       & 98.8 $\pm$ 0.1       & 96.6 $\pm$ 0.2       & 98.0                 \\
MLDG                 & 95.7 $\pm$ 0.2       & 98.9 $\pm$ 0.1       & 98.8 $\pm$ 0.1       & 98.9 $\pm$ 0.1       & 98.6 $\pm$ 0.1       & 95.8 $\pm$ 0.4       & 97.8                 \\
CORAL                & 96.2 $\pm$ 0.2       & 98.8 $\pm$ 0.1       & 98.8 $\pm$ 0.1       & 98.8 $\pm$ 0.1       & 98.9 $\pm$ 0.1       & 96.4 $\pm$ 0.2       & 98.0                 \\
MMD                  & 96.1 $\pm$ 0.2       & 98.9 $\pm$ 0.0       & 99.0 $\pm$ 0.0       & 98.8 $\pm$ 0.0       & 98.9 $\pm$ 0.0       & 96.4 $\pm$ 0.2       & 98.0                 \\
DANN                 & 95.9 $\pm$ 0.1       & 98.9 $\pm$ 0.1       & 98.6 $\pm$ 0.2       & 98.7 $\pm$ 0.1       & 98.9 $\pm$ 0.0       & 96.3 $\pm$ 0.3       & 97.9                 \\
CDANN                & 95.9 $\pm$ 0.2       & 98.8 $\pm$ 0.0       & 98.7 $\pm$ 0.1       & 98.9 $\pm$ 0.1       & 98.8 $\pm$ 0.1       & 96.1 $\pm$ 0.3       & 97.9                 \\
MTL                  & 96.1 $\pm$ 0.2       & 98.9 $\pm$ 0.0       & 99.0 $\pm$ 0.0       & 98.7 $\pm$ 0.1       & 99.0 $\pm$ 0.0       & 95.8 $\pm$ 0.3       & 97.9                 \\
SagNet               & 95.9 $\pm$ 0.1       & 99.0 $\pm$ 0.1       & 98.9 $\pm$ 0.1       & 98.6 $\pm$ 0.1       & 98.8 $\pm$ 0.1       & 96.3 $\pm$ 0.1       & 97.9                 \\
ARM                  & 95.9 $\pm$ 0.4       & 99.0 $\pm$ 0.1       & 98.8 $\pm$ 0.1       & 98.9 $\pm$ 0.1       & 99.1 $\pm$ 0.1       & 96.7 $\pm$ 0.2       & 98.1                 \\
VREx                 & 95.5 $\pm$ 0.2       & 99.0 $\pm$ 0.0       & 98.7 $\pm$ 0.2       & 98.8 $\pm$ 0.1       & 98.8 $\pm$ 0.0       & 96.4 $\pm$ 0.0       & 97.9                 \\
RSC                  & 95.4 $\pm$ 0.1       & 98.6 $\pm$ 0.1       & 98.6 $\pm$ 0.1       & 98.9 $\pm$ 0.0       & 98.8 $\pm$ 0.1       & 95.4 $\pm$ 0.3       & 97.6                 \\
\midrule
AND-mask              & 94.9 $\pm$ 0.1       & 98.8 $\pm$ 0.1       & 98.8 $\pm$ 0.1       & 98.7 $\pm$ 0.2       & 98.6 $\pm$ 0.2       & 95.5 $\pm$ 0.2       & 97.5                 \\
SAND-mask            & 94.7 $\pm$ 0.2       & 98.5 $\pm$ 0.2       & 98.6 $\pm$ 0.1       & 98.6 $\pm$ 0.1       & 98.5 $\pm$ 0.1       & 95.2 $\pm$ 0.1       & 97.4                 \\
\bottomrule
\end{tabular}}
\end{center}

\subsection{VLCS}

\begin{center}
\adjustbox{max width=\textwidth}{%
\begin{tabular}{lccccc}
\toprule
\multicolumn{6}{c}{\textbf{Model selection method: training domain validation set}} \\
\midrule
\textbf{Algorithm}   & \textbf{C}           & \textbf{L}           & \textbf{S}           & \textbf{V}           & \textbf{Avg}         \\
\midrule
ERM                  & 97.7 $\pm$ 0.4       & 64.3 $\pm$ 0.9       & 73.4 $\pm$ 0.5       & 74.6 $\pm$ 1.3       & 77.5                 \\
IRM                  & 98.6 $\pm$ 0.1       & 64.9 $\pm$ 0.9       & 73.4 $\pm$ 0.6       & 77.3 $\pm$ 0.9       & 78.5                 \\
GroupDRO             & 97.3 $\pm$ 0.3       & 63.4 $\pm$ 0.9       & 69.5 $\pm$ 0.8       & 76.7 $\pm$ 0.7       & 76.7                 \\
Mixup                & 98.3 $\pm$ 0.6       & 64.8 $\pm$ 1.0       & 72.1 $\pm$ 0.5       & 74.3 $\pm$ 0.8       & 77.4                 \\
MLDG                 & 97.4 $\pm$ 0.2       & 65.2 $\pm$ 0.7       & 71.0 $\pm$ 1.4       & 75.3 $\pm$ 1.0       & 77.2                 \\
CORAL                & 98.3 $\pm$ 0.1       & 66.1 $\pm$ 1.2       & 73.4 $\pm$ 0.3       & 77.5 $\pm$ 1.2       & 78.8                 \\
MMD                  & 97.7 $\pm$ 0.1       & 64.0 $\pm$ 1.1       & 72.8 $\pm$ 0.2       & 75.3 $\pm$ 3.3       & 77.5                 \\
DANN                 & 99.0 $\pm$ 0.3       & 65.1 $\pm$ 1.4       & 73.1 $\pm$ 0.3       & 77.2 $\pm$ 0.6       & 78.6                 \\
CDANN                & 97.1 $\pm$ 0.3       & 65.1 $\pm$ 1.2       & 70.7 $\pm$ 0.8       & 77.1 $\pm$ 1.5       & 77.5                 \\
MTL                  & 97.8 $\pm$ 0.4       & 64.3 $\pm$ 0.3       & 71.5 $\pm$ 0.7       & 75.3 $\pm$ 1.7       & 77.2                 \\
SagNet               & 97.9 $\pm$ 0.4       & 64.5 $\pm$ 0.5       & 71.4 $\pm$ 1.3       & 77.5 $\pm$ 0.5       & 77.8                 \\
ARM                  & 98.7 $\pm$ 0.2       & 63.6 $\pm$ 0.7       & 71.3 $\pm$ 1.2       & 76.7 $\pm$ 0.6       & 77.6                 \\
VREx                 & 98.4 $\pm$ 0.3       & 64.4 $\pm$ 1.4       & 74.1 $\pm$ 0.4       & 76.2 $\pm$ 1.3       & 78.3                 \\
RSC                  & 97.9 $\pm$ 0.1       & 62.5 $\pm$ 0.7       & 72.3 $\pm$ 1.2       & 75.6 $\pm$ 0.8       & 77.1                 \\
\midrule
AND-mask              & 97.8 $\pm$ 0.4       & 64.3 $\pm$ 1.2       & 73.5 $\pm$ 0.7       & 76.8 $\pm$ 2.6       & 78.1                 \\
SAND-mask            & 98.5 $\pm$ 0.3       & 63.6 $\pm$ 0.9       & 70.4 $\pm$ 0.8       & 77.1 $\pm$ 0.8       & 77.4                 \\

\bottomrule
\end{tabular}}
\end{center}

\begin{center}
\adjustbox{max width=\textwidth}{%
\begin{tabular}{lccccc}
\toprule
\multicolumn{6}{c}{\textbf{Model selection method: test-domain validation set \textit{(oracle)}}} \\
\midrule
\textbf{Algorithm}   & \textbf{C}           & \textbf{L}           & \textbf{S}           & \textbf{V}           & \textbf{Avg}         \\
\midrule
ERM                  & 97.6 $\pm$ 0.3       & 67.9 $\pm$ 0.7       & 70.9 $\pm$ 0.2       & 74.0 $\pm$ 0.6       & 77.6                 \\
IRM                  & 97.3 $\pm$ 0.2       & 66.7 $\pm$ 0.1       & 71.0 $\pm$ 2.3       & 72.8 $\pm$ 0.4       & 76.9                 \\
GroupDRO             & 97.7 $\pm$ 0.2       & 65.9 $\pm$ 0.2       & 72.8 $\pm$ 0.8       & 73.4 $\pm$ 1.3       & 77.4                 \\
Mixup                & 97.8 $\pm$ 0.4       & 67.2 $\pm$ 0.4       & 71.5 $\pm$ 0.2       & 75.7 $\pm$ 0.6       & 78.1                 \\
MLDG                 & 97.1 $\pm$ 0.5       & 66.6 $\pm$ 0.5       & 71.5 $\pm$ 0.1       & 75.0 $\pm$ 0.9       & 77.5                 \\
CORAL                & 97.3 $\pm$ 0.2       & 67.5 $\pm$ 0.6       & 71.6 $\pm$ 0.6       & 74.5 $\pm$ 0.0       & 77.7                 \\
MMD                  & 98.8 $\pm$ 0.0       & 66.4 $\pm$ 0.4       & 70.8 $\pm$ 0.5       & 75.6 $\pm$ 0.4       & 77.9                 \\
DANN                 & 99.0 $\pm$ 0.2       & 66.3 $\pm$ 1.2       & 73.4 $\pm$ 1.4       & 80.1 $\pm$ 0.5       & 79.7                 \\
CDANN                & 98.2 $\pm$ 0.1       & 68.8 $\pm$ 0.5       & 74.3 $\pm$ 0.6       & 78.1 $\pm$ 0.5       & 79.9                 \\
MTL                  & 97.9 $\pm$ 0.7       & 66.1 $\pm$ 0.7       & 72.0 $\pm$ 0.4       & 74.9 $\pm$ 1.1       & 77.7                 \\
SagNet               & 97.4 $\pm$ 0.3       & 66.4 $\pm$ 0.4       & 71.6 $\pm$ 0.1       & 75.0 $\pm$ 0.8       & 77.6                 \\
ARM                  & 97.6 $\pm$ 0.6       & 66.5 $\pm$ 0.3       & 72.7 $\pm$ 0.6       & 74.4 $\pm$ 0.7       & 77.8                 \\
VREx                 & 98.4 $\pm$ 0.2       & 66.4 $\pm$ 0.7       & 72.8 $\pm$ 0.1       & 75.0 $\pm$ 1.4       & 78.1                 \\
RSC                  & 98.0 $\pm$ 0.4       & 67.2 $\pm$ 0.3       & 70.3 $\pm$ 1.3       & 75.6 $\pm$ 0.4       & 77.8                 \\
\midrule
AND-mask              & 98.3 $\pm$ 0.3       & 64.5 $\pm$ 0.2       & 69.3 $\pm$ 1.3       & 73.4 $\pm$ 1.3       & 76.4                 \\
SAND-mask            & 97.6 $\pm$ 0.3       & 64.5 $\pm$ 0.6       & 69.7 $\pm$ 0.6       & 73.0 $\pm$ 1.2       & 76.2                 \\
\bottomrule
\end{tabular}}
\end{center}

\subsection{PACS}

\begin{center}
\adjustbox{max width=\textwidth}{%
\begin{tabular}{lccccc}
\toprule
\multicolumn{6}{c}{\textbf{Model selection method: training domain validation set}} \\
\midrule
\textbf{Algorithm}   & \textbf{A}           & \textbf{C}           & \textbf{P}           & \textbf{S}           & \textbf{Avg}         \\
\midrule
ERM                  & 84.7 $\pm$ 0.4       & 80.8 $\pm$ 0.6       & 97.2 $\pm$ 0.3       & 79.3 $\pm$ 1.0       & 85.5                 \\
IRM                  & 84.8 $\pm$ 1.3       & 76.4 $\pm$ 1.1       & 96.7 $\pm$ 0.6       & 76.1 $\pm$ 1.0       & 83.5                 \\
GroupDRO             & 83.5 $\pm$ 0.9       & 79.1 $\pm$ 0.6       & 96.7 $\pm$ 0.3       & 78.3 $\pm$ 2.0       & 84.4                 \\
Mixup                & 86.1 $\pm$ 0.5       & 78.9 $\pm$ 0.8       & 97.6 $\pm$ 0.1       & 75.8 $\pm$ 1.8       & 84.6                 \\
MLDG                 & 85.5 $\pm$ 1.4       & 80.1 $\pm$ 1.7       & 97.4 $\pm$ 0.3       & 76.6 $\pm$ 1.1       & 84.9                 \\
CORAL                & 88.3 $\pm$ 0.2       & 80.0 $\pm$ 0.5       & 97.5 $\pm$ 0.3       & 78.8 $\pm$ 1.3       & 86.2                 \\
MMD                  & 86.1 $\pm$ 1.4       & 79.4 $\pm$ 0.9       & 96.6 $\pm$ 0.2       & 76.5 $\pm$ 0.5       & 84.6                 \\
DANN                 & 86.4 $\pm$ 0.8       & 77.4 $\pm$ 0.8       & 97.3 $\pm$ 0.4       & 73.5 $\pm$ 2.3       & 83.6                 \\
CDANN                & 84.6 $\pm$ 1.8       & 75.5 $\pm$ 0.9       & 96.8 $\pm$ 0.3       & 73.5 $\pm$ 0.6       & 82.6                 \\
MTL                  & 87.5 $\pm$ 0.8       & 77.1 $\pm$ 0.5       & 96.4 $\pm$ 0.8       & 77.3 $\pm$ 1.8       & 84.6                 \\
SagNet               & 87.4 $\pm$ 1.0       & 80.7 $\pm$ 0.6       & 97.1 $\pm$ 0.1       & 80.0 $\pm$ 0.4       & 86.3                 \\
ARM                  & 86.8 $\pm$ 0.6       & 76.8 $\pm$ 0.5       & 97.4 $\pm$ 0.3       & 79.3 $\pm$ 1.2       & 85.1                 \\
VREx                 & 86.0 $\pm$ 1.6       & 79.1 $\pm$ 0.6       & 96.9 $\pm$ 0.5       & 77.7 $\pm$ 1.7       & 84.9                 \\
RSC                  & 85.4 $\pm$ 0.8       & 79.7 $\pm$ 1.8       & 97.6 $\pm$ 0.3       & 78.2 $\pm$ 1.2       & 85.2                 \\
\midrule
AND-mask              & 85.3 $\pm$ 1.4       & 79.2 $\pm$ 2.0       & 96.9 $\pm$ 0.4       & 76.2 $\pm$ 1.4       & 84.4                 \\
SAND-mask            & 85.8 $\pm$ 1.7       & 79.2 $\pm$ 0.8       & 96.3 $\pm$ 0.2       & 76.9 $\pm$ 2.0       & 84.6                 \\
\bottomrule
\end{tabular}}
\end{center}

\begin{center}
\adjustbox{max width=\textwidth}{%
\begin{tabular}{lccccc}
\toprule
\multicolumn{6}{c}{\textbf{Model selection method: test-domain validation set \textit{(oracle)}}} \\
\midrule
\textbf{Algorithm}   & \textbf{A}           & \textbf{C}           & \textbf{P}           & \textbf{S}           & \textbf{Avg}         \\
\midrule
ERM                  & 86.5 $\pm$ 1.0       & 81.3 $\pm$ 0.6       & 96.2 $\pm$ 0.3       & 82.7 $\pm$ 1.1       & 86.7                 \\
IRM                  & 84.2 $\pm$ 0.9       & 79.7 $\pm$ 1.5       & 95.9 $\pm$ 0.4       & 78.3 $\pm$ 2.1       & 84.5                 \\
GroupDRO             & 87.5 $\pm$ 0.5       & 82.9 $\pm$ 0.6       & 97.1 $\pm$ 0.3       & 81.1 $\pm$ 1.2       & 87.1                 \\
Mixup                & 87.5 $\pm$ 0.4       & 81.6 $\pm$ 0.7       & 97.4 $\pm$ 0.2       & 80.8 $\pm$ 0.9       & 86.8                 \\
MLDG                 & 87.0 $\pm$ 1.2       & 82.5 $\pm$ 0.9       & 96.7 $\pm$ 0.3       & 81.2 $\pm$ 0.6       & 86.8                 \\
CORAL                & 86.6 $\pm$ 0.8       & 81.8 $\pm$ 0.9       & 97.1 $\pm$ 0.5       & 82.7 $\pm$ 0.6       & 87.1                 \\
MMD                  & 88.1 $\pm$ 0.8       & 82.6 $\pm$ 0.7       & 97.1 $\pm$ 0.5       & 81.2 $\pm$ 1.2       & 87.2                 \\
DANN                 & 87.0 $\pm$ 0.4       & 80.3 $\pm$ 0.6       & 96.8 $\pm$ 0.3       & 76.9 $\pm$ 1.1       & 85.2                 \\
CDANN                & 87.7 $\pm$ 0.6       & 80.7 $\pm$ 1.2       & 97.3 $\pm$ 0.4       & 77.6 $\pm$ 1.5       & 85.8                 \\
MTL                  & 87.0 $\pm$ 0.2       & 82.7 $\pm$ 0.8       & 96.5 $\pm$ 0.7       & 80.5 $\pm$ 0.8       & 86.7                 \\
SagNet               & 87.4 $\pm$ 0.5       & 81.2 $\pm$ 1.2       & 96.3 $\pm$ 0.8       & 80.7 $\pm$ 1.1       & 86.4                 \\
ARM                  & 85.0 $\pm$ 1.2       & 81.4 $\pm$ 0.2       & 95.9 $\pm$ 0.3       & 80.9 $\pm$ 0.5       & 85.8                 \\
VREx                 & 87.8 $\pm$ 1.2       & 81.8 $\pm$ 0.7       & 97.4 $\pm$ 0.2       & 82.1 $\pm$ 0.7       & 87.2                 \\
RSC                  & 86.0 $\pm$ 0.7       & 81.8 $\pm$ 0.9       & 96.8 $\pm$ 0.7       & 80.4 $\pm$ 0.5       & 86.2                 \\
\midrule
AND-mask              & 86.4 $\pm$ 1.1       & 80.8 $\pm$ 0.9       & 97.1 $\pm$ 0.2       & 81.3 $\pm$ 1.1       & 86.4                 \\
SAND-mask            & 86.1 $\pm$ 0.6       & 80.3 $\pm$ 1.0       & 97.1 $\pm$ 0.3       & 80.0 $\pm$ 1.3       & 85.9                 \\
\bottomrule
\end{tabular}}
\end{center}

\subsection{OfficeHome}

\begin{center}
\adjustbox{max width=\textwidth}{%
\begin{tabular}{lccccc}
\toprule
\multicolumn{6}{c}{\textbf{Model selection method: training domain validation set}} \\
\midrule
\textbf{Algorithm}   & \textbf{A}           & \textbf{C}           & \textbf{P}           & \textbf{R}           & \textbf{Avg}         \\
\midrule
ERM                  & 61.3 $\pm$ 0.7       & 52.4 $\pm$ 0.3       & 75.8 $\pm$ 0.1       & 76.6 $\pm$ 0.3       & 66.5                 \\
IRM                  & 58.9 $\pm$ 2.3       & 52.2 $\pm$ 1.6       & 72.1 $\pm$ 2.9       & 74.0 $\pm$ 2.5       & 64.3                 \\
GroupDRO             & 60.4 $\pm$ 0.7       & 52.7 $\pm$ 1.0       & 75.0 $\pm$ 0.7       & 76.0 $\pm$ 0.7       & 66.0                 \\
Mixup                & 62.4 $\pm$ 0.8       & 54.8 $\pm$ 0.6       & 76.9 $\pm$ 0.3       & 78.3 $\pm$ 0.2       & 68.1                 \\
MLDG                 & 61.5 $\pm$ 0.9       & 53.2 $\pm$ 0.6       & 75.0 $\pm$ 1.2       & 77.5 $\pm$ 0.4       & 66.8                 \\
CORAL                & 65.3 $\pm$ 0.4       & 54.4 $\pm$ 0.5       & 76.5 $\pm$ 0.1       & 78.4 $\pm$ 0.5       & 68.7                 \\
MMD                  & 60.4 $\pm$ 0.2       & 53.3 $\pm$ 0.3       & 74.3 $\pm$ 0.1       & 77.4 $\pm$ 0.6       & 66.3                 \\
DANN                 & 59.9 $\pm$ 1.3       & 53.0 $\pm$ 0.3       & 73.6 $\pm$ 0.7       & 76.9 $\pm$ 0.5       & 65.9                 \\
CDANN                & 61.5 $\pm$ 1.4       & 50.4 $\pm$ 2.4       & 74.4 $\pm$ 0.9       & 76.6 $\pm$ 0.8       & 65.8                 \\
MTL                  & 61.5 $\pm$ 0.7       & 52.4 $\pm$ 0.6       & 74.9 $\pm$ 0.4       & 76.8 $\pm$ 0.4       & 66.4                 \\
SagNet               & 63.4 $\pm$ 0.2       & 54.8 $\pm$ 0.4       & 75.8 $\pm$ 0.4       & 78.3 $\pm$ 0.3       & 68.1                 \\
ARM                  & 58.9 $\pm$ 0.8       & 51.0 $\pm$ 0.5       & 74.1 $\pm$ 0.1       & 75.2 $\pm$ 0.3       & 64.8                 \\
VREx                 & 60.7 $\pm$ 0.9       & 53.0 $\pm$ 0.9       & 75.3 $\pm$ 0.1       & 76.6 $\pm$ 0.5       & 66.4                 \\
RSC                  & 60.7 $\pm$ 1.4       & 51.4 $\pm$ 0.3       & 74.8 $\pm$ 1.1       & 75.1 $\pm$ 1.3       & 65.5                 \\
\midrule
ANDMask              & 59.5 $\pm$ 1.2       & 51.7 $\pm$ 0.2       & 73.9 $\pm$ 0.4       & 77.1 $\pm$ 0.2       & 65.6                 \\
SAND-mask            & 60.3 $\pm$ 0.5       & 53.3 $\pm$ 0.7       & 73.5 $\pm$ 0.7       & 76.2 $\pm$ 0.3       & 65.8                 \\
\bottomrule
\end{tabular}}
\end{center}

\begin{center}
\adjustbox{max width=\textwidth}{%
\begin{tabular}{lccccc}
\toprule
\multicolumn{6}{c}{\textbf{Model selection method: test-domain validation set \textit{(oracle)}}} \\
\midrule
\textbf{Algorithm}   & \textbf{A}           & \textbf{C}           & \textbf{P}           & \textbf{R}           & \textbf{Avg}         \\
\midrule
ERM                  & 61.7 $\pm$ 0.7       & 53.4 $\pm$ 0.3       & 74.1 $\pm$ 0.4       & 76.2 $\pm$ 0.6       & 66.4                 \\
IRM                  & 56.4 $\pm$ 3.2       & 51.2 $\pm$ 2.3       & 71.7 $\pm$ 2.7       & 72.7 $\pm$ 2.7       & 63.0                 \\
GroupDRO             & 60.5 $\pm$ 1.6       & 53.1 $\pm$ 0.3       & 75.5 $\pm$ 0.3       & 75.9 $\pm$ 0.7       & 66.2                 \\
Mixup                & 63.5 $\pm$ 0.2       & 54.6 $\pm$ 0.4       & 76.0 $\pm$ 0.3       & 78.0 $\pm$ 0.7       & 68.0                 \\
MLDG                 & 60.5 $\pm$ 0.7       & 54.2 $\pm$ 0.5       & 75.0 $\pm$ 0.2       & 76.7 $\pm$ 0.5       & 66.6                 \\
CORAL                & 64.8 $\pm$ 0.8       & 54.1 $\pm$ 0.9       & 76.5 $\pm$ 0.4       & 78.2 $\pm$ 0.4       & 68.4                 \\
MMD                  & 60.4 $\pm$ 1.0       & 53.4 $\pm$ 0.5       & 74.9 $\pm$ 0.1       & 76.1 $\pm$ 0.7       & 66.2                 \\
DANN                 & 60.6 $\pm$ 1.4       & 51.8 $\pm$ 0.7       & 73.4 $\pm$ 0.5       & 75.5 $\pm$ 0.9       & 65.3                 \\
CDANN                & 57.9 $\pm$ 0.2       & 52.1 $\pm$ 1.2       & 74.9 $\pm$ 0.7       & 76.2 $\pm$ 0.2       & 65.3                 \\
MTL                  & 60.7 $\pm$ 0.8       & 53.5 $\pm$ 1.3       & 75.2 $\pm$ 0.6       & 76.6 $\pm$ 0.6       & 66.5                 \\
SagNet               & 62.7 $\pm$ 0.5       & 53.6 $\pm$ 0.5       & 76.0 $\pm$ 0.3       & 77.8 $\pm$ 0.1       & 67.5                 \\
ARM                  & 58.8 $\pm$ 0.5       & 51.8 $\pm$ 0.7       & 74.0 $\pm$ 0.1       & 74.4 $\pm$ 0.2       & 64.8                 \\
VREx                 & 59.6 $\pm$ 1.0       & 53.3 $\pm$ 0.3       & 73.2 $\pm$ 0.5       & 76.6 $\pm$ 0.4       & 65.7                 \\
RSC                  & 61.7 $\pm$ 0.8       & 53.0 $\pm$ 0.9       & 74.8 $\pm$ 0.8       & 76.3 $\pm$ 0.5       & 66.5                 \\
\midrule
ANDMask              & 60.3 $\pm$ 0.5       & 52.3 $\pm$ 0.6       & 75.1 $\pm$ 0.2       & 76.6 $\pm$ 0.3       & 66.1                 \\
SAND-mask            & 59.9 $\pm$ 0.7       & 53.6 $\pm$ 0.8       & 74.3 $\pm$ 0.4       & 75.8 $\pm$ 0.5       & 65.9                 \\
\bottomrule
\end{tabular}}
\end{center}

\subsection{TerraIncognita}

\begin{center}
\adjustbox{max width=\textwidth}{%
\begin{tabular}{lccccc}
\toprule
\multicolumn{6}{c}{\textbf{Model selection method: training domain validation set}} \\
\midrule
\textbf{Algorithm}   & \textbf{L100}        & \textbf{L38}         & \textbf{L43}         & \textbf{L46}         & \textbf{Avg}         \\
\midrule
ERM                  & 49.8 $\pm$ 4.4       & 42.1 $\pm$ 1.4       & 56.9 $\pm$ 1.8       & 35.7 $\pm$ 3.9       & 46.1                 \\
IRM                  & 54.6 $\pm$ 1.3       & 39.8 $\pm$ 1.9       & 56.2 $\pm$ 1.8       & 39.6 $\pm$ 0.8       & 47.6                 \\
GroupDRO             & 41.2 $\pm$ 0.7       & 38.6 $\pm$ 2.1       & 56.7 $\pm$ 0.9       & 36.4 $\pm$ 2.1       & 43.2                 \\
Mixup                & 59.6 $\pm$ 2.0       & 42.2 $\pm$ 1.4       & 55.9 $\pm$ 0.8       & 33.9 $\pm$ 1.4       & 47.9                 \\
MLDG                 & 54.2 $\pm$ 3.0       & 44.3 $\pm$ 1.1       & 55.6 $\pm$ 0.3       & 36.9 $\pm$ 2.2       & 47.7                 \\
CORAL                & 51.6 $\pm$ 2.4       & 42.2 $\pm$ 1.0       & 57.0 $\pm$ 1.0       & 39.8 $\pm$ 2.9       & 47.6                 \\
MMD                  & 41.9 $\pm$ 3.0       & 34.8 $\pm$ 1.0       & 57.0 $\pm$ 1.9       & 35.2 $\pm$ 1.8       & 42.2                 \\
DANN                 & 51.1 $\pm$ 3.5       & 40.6 $\pm$ 0.6       & 57.4 $\pm$ 0.5       & 37.7 $\pm$ 1.8       & 46.7                 \\
CDANN                & 47.0 $\pm$ 1.9       & 41.3 $\pm$ 4.8       & 54.9 $\pm$ 1.7       & 39.8 $\pm$ 2.3       & 45.8                 \\
MTL                  & 49.3 $\pm$ 1.2       & 39.6 $\pm$ 6.3       & 55.6 $\pm$ 1.1       & 37.8 $\pm$ 0.8       & 45.6                 \\
SagNet               & 53.0 $\pm$ 2.9       & 43.0 $\pm$ 2.5       & 57.9 $\pm$ 0.6       & 40.4 $\pm$ 1.3       & 48.6                 \\
ARM                  & 49.3 $\pm$ 0.7       & 38.3 $\pm$ 2.4       & 55.8 $\pm$ 0.8       & 38.7 $\pm$ 1.3       & 45.5                 \\
VREx                 & 48.2 $\pm$ 4.3       & 41.7 $\pm$ 1.3       & 56.8 $\pm$ 0.8       & 38.7 $\pm$ 3.1       & 46.4                 \\
RSC                  & 50.2 $\pm$ 2.2       & 39.2 $\pm$ 1.4       & 56.3 $\pm$ 1.4       & 40.8 $\pm$ 0.6       & 46.6                 \\
\midrule
AND-mask              & 50.0 $\pm$ 2.9       & 40.2 $\pm$ 0.8       & 53.3 $\pm$ 0.7       & 34.8 $\pm$ 1.9       & 44.6                 \\
SAND-mask            & 45.7 $\pm$ 2.9       & 31.6 $\pm$ 4.7       & 55.1 $\pm$ 1.0       & 39.0 $\pm$ 1.8       & 42.9                 \\

\bottomrule
\end{tabular}}
\end{center}

\begin{center}
\adjustbox{max width=\textwidth}{%
\begin{tabular}{lccccc}
\toprule
\multicolumn{6}{c}{\textbf{Model selection method: test-domain validation set \textit{(oracle)}}} \\
\midrule
\textbf{Algorithm}   & \textbf{L100}        & \textbf{L38}         & \textbf{L43}         & \textbf{L46}         & \textbf{Avg}         \\
\midrule
ERM                  & 59.4 $\pm$ 0.9       & 49.3 $\pm$ 0.6       & 60.1 $\pm$ 1.1       & 43.2 $\pm$ 0.5       & 53.0                 \\
IRM                  & 56.5 $\pm$ 2.5       & 49.8 $\pm$ 1.5       & 57.1 $\pm$ 2.2       & 38.6 $\pm$ 1.0       & 50.5                 \\
GroupDRO             & 60.4 $\pm$ 1.5       & 48.3 $\pm$ 0.4       & 58.6 $\pm$ 0.8       & 42.2 $\pm$ 0.8       & 52.4                 \\
Mixup                & 67.6 $\pm$ 1.8       & 51.0 $\pm$ 1.3       & 59.0 $\pm$ 0.0       & 40.0 $\pm$ 1.1       & 54.4                 \\
MLDG                 & 59.2 $\pm$ 0.1       & 49.0 $\pm$ 0.9       & 58.4 $\pm$ 0.9       & 41.4 $\pm$ 1.0       & 52.0                 \\
CORAL                & 60.4 $\pm$ 0.9       & 47.2 $\pm$ 0.5       & 59.3 $\pm$ 0.4       & 44.4 $\pm$ 0.4       & 52.8                 \\
MMD                  & 60.6 $\pm$ 1.1       & 45.9 $\pm$ 0.3       & 57.8 $\pm$ 0.5       & 43.8 $\pm$ 1.2       & 52.0                 \\
DANN                 & 55.2 $\pm$ 1.9       & 47.0 $\pm$ 0.7       & 57.2 $\pm$ 0.9       & 42.9 $\pm$ 0.9       & 50.6                 \\
CDANN                & 56.3 $\pm$ 2.0       & 47.1 $\pm$ 0.9       & 57.2 $\pm$ 1.1       & 42.4 $\pm$ 0.8       & 50.8                 \\
MTL                  & 58.4 $\pm$ 2.1       & 48.4 $\pm$ 0.8       & 58.9 $\pm$ 0.6       & 43.0 $\pm$ 1.3       & 52.2                 \\
SagNet               & 56.4 $\pm$ 1.9       & 50.5 $\pm$ 2.3       & 59.1 $\pm$ 0.5       & 44.1 $\pm$ 0.6       & 52.5                 \\
ARM                  & 60.1 $\pm$ 1.5       & 48.3 $\pm$ 1.6       & 55.3 $\pm$ 0.6       & 40.9 $\pm$ 1.1       & 51.2                 \\
VREx                 & 56.8 $\pm$ 1.7       & 46.5 $\pm$ 0.5       & 58.4 $\pm$ 0.3       & 43.8 $\pm$ 0.3       & 51.4                 \\
RSC                  & 59.9 $\pm$ 1.4       & 46.7 $\pm$ 0.4       & 57.8 $\pm$ 0.5       & 44.3 $\pm$ 0.6       & 52.1                 \\
\midrule
AND-mask              & 54.7 $\pm$ 1.8       & 48.4 $\pm$ 0.5       & 55.1 $\pm$ 0.5       & 41.3 $\pm$ 0.6       & 49.8                 \\
SAND-mask            & 56.2 $\pm$ 1.8       & 46.3 $\pm$ 0.3       & 55.8 $\pm$ 0.4       & 42.6 $\pm$ 1.2       & 50.2                 \\
\bottomrule
\end{tabular}}
\end{center}

\subsubsection{DomainNet}

\begin{center}
\adjustbox{max width=\textwidth}{%
\begin{tabular}{lccccccc}
\toprule
\multicolumn{8}{c}{\textbf{Model selection method: training domain validation set}} \\
\midrule
\textbf{Algorithm}   & \textbf{clip}        & \textbf{info}        & \textbf{paint}       & \textbf{quick}       & \textbf{real}        & \textbf{sketch}      & \textbf{Avg}         \\
\midrule
ERM                  & 58.1 $\pm$ 0.3       & 18.8 $\pm$ 0.3       & 46.7 $\pm$ 0.3       & 12.2 $\pm$ 0.4       & 59.6 $\pm$ 0.1       & 49.8 $\pm$ 0.4       & 40.9                 \\
IRM                  & 48.5 $\pm$ 2.8       & 15.0 $\pm$ 1.5       & 38.3 $\pm$ 4.3       & 10.9 $\pm$ 0.5       & 48.2 $\pm$ 5.2       & 42.3 $\pm$ 3.1       & 33.9                 \\
GroupDRO             & 47.2 $\pm$ 0.5       & 17.5 $\pm$ 0.4       & 33.8 $\pm$ 0.5       & 9.3 $\pm$ 0.3        & 51.6 $\pm$ 0.4       & 40.1 $\pm$ 0.6       & 33.3                 \\
Mixup                & 55.7 $\pm$ 0.3       & 18.5 $\pm$ 0.5       & 44.3 $\pm$ 0.5       & 12.5 $\pm$ 0.4       & 55.8 $\pm$ 0.3       & 48.2 $\pm$ 0.5       & 39.2                 \\
MLDG                 & 59.1 $\pm$ 0.2       & 19.1 $\pm$ 0.3       & 45.8 $\pm$ 0.7       & 13.4 $\pm$ 0.3       & 59.6 $\pm$ 0.2       & 50.2 $\pm$ 0.4       & 41.2                 \\
CORAL                & 59.2 $\pm$ 0.1       & 19.7 $\pm$ 0.2       & 46.6 $\pm$ 0.3       & 13.4 $\pm$ 0.4       & 59.8 $\pm$ 0.2       & 50.1 $\pm$ 0.6       & 41.5                 \\
MMD                  & 32.1 $\pm$ 13.3      & 11.0 $\pm$ 4.6       & 26.8 $\pm$ 11.3      & 8.7 $\pm$ 2.1        & 32.7 $\pm$ 13.8      & 28.9 $\pm$ 11.9      & 23.4                 \\
DANN                 & 53.1 $\pm$ 0.2       & 18.3 $\pm$ 0.1       & 44.2 $\pm$ 0.7       & 11.8 $\pm$ 0.1       & 55.5 $\pm$ 0.4       & 46.8 $\pm$ 0.6       & 38.3                 \\
CDANN                & 54.6 $\pm$ 0.4       & 17.3 $\pm$ 0.1       & 43.7 $\pm$ 0.9       & 12.1 $\pm$ 0.7       & 56.2 $\pm$ 0.4       & 45.9 $\pm$ 0.5       & 38.3                 \\
MTL                  & 57.9 $\pm$ 0.5       & 18.5 $\pm$ 0.4       & 46.0 $\pm$ 0.1       & 12.5 $\pm$ 0.1       & 59.5 $\pm$ 0.3       & 49.2 $\pm$ 0.1       & 40.6                 \\
SagNet               & 57.7 $\pm$ 0.3       & 19.0 $\pm$ 0.2       & 45.3 $\pm$ 0.3       & 12.7 $\pm$ 0.5       & 58.1 $\pm$ 0.5       & 48.8 $\pm$ 0.2       & 40.3                 \\
ARM                  & 49.7 $\pm$ 0.3       & 16.3 $\pm$ 0.5       & 40.9 $\pm$ 1.1       & 9.4 $\pm$ 0.1        & 53.4 $\pm$ 0.4       & 43.5 $\pm$ 0.4       & 35.5                 \\
VREx                 & 47.3 $\pm$ 3.5       & 16.0 $\pm$ 1.5       & 35.8 $\pm$ 4.6       & 10.9 $\pm$ 0.3       & 49.6 $\pm$ 4.9       & 42.0 $\pm$ 3.0       & 33.6                 \\
RSC                  & 55.0 $\pm$ 1.2       & 18.3 $\pm$ 0.5       & 44.4 $\pm$ 0.6       & 12.2 $\pm$ 0.2       & 55.7 $\pm$ 0.7       & 47.8 $\pm$ 0.9       & 38.9                 \\
\midrule
ANDMask              & 52.3 $\pm$ 0.8       & 16.6 $\pm$ 0.3       & 41.6 $\pm$ 1.1       & 11.3 $\pm$ 0.1       & 55.8 $\pm$ 0.4       & 45.4 $\pm$ 0.9       & 37.2                 \\
SANDMask             & 43.8 $\pm$ 1.3       & 14.8 $\pm$ 0.3       & 38.2 $\pm$ 0.6       & 9.0 $\pm$ 0.3        & 47.0 $\pm$ 1.1       & 39.9 $\pm$ 0.6       & 32.1                 \\
\bottomrule
\end{tabular}}
\end{center}

\begin{center}
\adjustbox{max width=\textwidth}{%
\begin{tabular}{lccccccc}
\toprule
\multicolumn{8}{c}{\textbf{Model selection method: test-domain validation set \textit{(oracle)}}} \\
\midrule
\textbf{Algorithm}   & \textbf{clip}        & \textbf{info}        & \textbf{paint}       & \textbf{quick}       & \textbf{real}        & \textbf{sketch}      & \textbf{Avg}         \\
\midrule
ERM                  & 58.6 $\pm$ 0.3       & 19.2 $\pm$ 0.2       & 47.0 $\pm$ 0.3       & 13.2 $\pm$ 0.2       & 59.9 $\pm$ 0.3       & 49.8 $\pm$ 0.4       & 41.3                 \\
IRM                  & 40.4 $\pm$ 6.6       & 12.1 $\pm$ 2.7       & 31.4 $\pm$ 5.7       & 9.8 $\pm$ 1.2        & 37.7 $\pm$ 9.0       & 36.7 $\pm$ 5.3       & 28.0                 \\
GroupDRO             & 47.2 $\pm$ 0.5       & 17.5 $\pm$ 0.4       & 34.2 $\pm$ 0.3       & 9.2 $\pm$ 0.4        & 51.9 $\pm$ 0.5       & 40.1 $\pm$ 0.6       & 33.4                 \\
Mixup                & 55.6 $\pm$ 0.1       & 18.7 $\pm$ 0.4       & 45.1 $\pm$ 0.5       & 12.8 $\pm$ 0.3       & 57.6 $\pm$ 0.5       & 48.2 $\pm$ 0.4       & 39.6                 \\
MLDG                 & 59.3 $\pm$ 0.1       & 19.6 $\pm$ 0.2       & 46.8 $\pm$ 0.2       & 13.4 $\pm$ 0.2       & 60.1 $\pm$ 0.4       & 50.4 $\pm$ 0.3       & 41.6                 \\
CORAL                & 59.2 $\pm$ 0.1       & 19.9 $\pm$ 0.2       & 47.4 $\pm$ 0.2       & 14.0 $\pm$ 0.4       & 59.8 $\pm$ 0.2       & 50.4 $\pm$ 0.4       & 41.8                 \\
MMD                  & 32.2 $\pm$ 13.3      & 11.2 $\pm$ 4.5       & 26.8 $\pm$ 11.3      & 8.8 $\pm$ 2.2        & 32.7 $\pm$ 13.8      & 29.0 $\pm$ 11.8      & 23.5                 \\
DANN                 & 53.1 $\pm$ 0.2       & 18.3 $\pm$ 0.1       & 44.2 $\pm$ 0.7       & 11.9 $\pm$ 0.1       & 55.5 $\pm$ 0.4       & 46.8 $\pm$ 0.6       & 38.3                 \\
CDANN                & 54.6 $\pm$ 0.4       & 17.3 $\pm$ 0.1       & 44.2 $\pm$ 0.7       & 12.8 $\pm$ 0.2       & 56.2 $\pm$ 0.4       & 45.9 $\pm$ 0.5       & 38.5                 \\
MTL                  & 58.0 $\pm$ 0.4       & 19.2 $\pm$ 0.2       & 46.2 $\pm$ 0.1       & 12.7 $\pm$ 0.2       & 59.9 $\pm$ 0.1       & 49.0 $\pm$ 0.0       & 40.8                 \\
SagNet               & 57.7 $\pm$ 0.3       & 19.1 $\pm$ 0.1       & 46.3 $\pm$ 0.5       & 13.5 $\pm$ 0.4       & 58.9 $\pm$ 0.4       & 49.5 $\pm$ 0.2       & 40.8                 \\
ARM                  & 49.6 $\pm$ 0.4       & 16.5 $\pm$ 0.3       & 41.5 $\pm$ 0.8       & 10.8 $\pm$ 0.1       & 53.5 $\pm$ 0.3       & 43.9 $\pm$ 0.4       & 36.0                 \\
VREx                 & 43.3 $\pm$ 4.5       & 14.1 $\pm$ 1.8       & 32.5 $\pm$ 5.0       & 9.8 $\pm$ 1.1        & 43.5 $\pm$ 5.6       & 37.7 $\pm$ 4.5       & 30.1                 \\
RSC                  & 55.0 $\pm$ 1.2       & 18.3 $\pm$ 0.5       & 44.4 $\pm$ 0.6       & 12.5 $\pm$ 0.1       & 55.7 $\pm$ 0.7       & 47.8 $\pm$ 0.9       & 38.9                 \\
\midrule
ANDMask              & 52.3 $\pm$ 0.8       & 17.3 $\pm$ 0.5       & 43.7 $\pm$ 1.1       & 12.3 $\pm$ 0.4       & 55.8 $\pm$ 0.4       & 46.1 $\pm$ 0.8       & 37.9                 \\
SANDMask             & 43.8 $\pm$ 1.3       & 15.2 $\pm$ 0.2       & 38.2 $\pm$ 0.6       & 9.0 $\pm$ 0.2        & 47.1 $\pm$ 1.1       & 39.9 $\pm$ 0.6       & 32.2                 \\

\bottomrule
\end{tabular}}
\end{center}

\end{document}

%% file: math_commands.tex
\usepackage{amsmath,amsfonts,bm}

\def\eqref#1{equation~\ref{#1}}

\def\1{\bm{1}}

\DeclareMathAlphabet{\mathsfit}{\encodingdefault}{\sfdefault}{m}{sl}
\SetMathAlphabet{\mathsfit}{bold}{\encodingdefault}{\sfdefault}{bx}{n}

